\documentclass[letterpaper,twocolumn,10pt]{article}
\usepackage{zhanggroup}

\usepackage{tikz}
\usepackage{amsmath}
\usepackage{caption}
\usepackage{subcaption}
\usepackage{float}
\usepackage{multirow}
\usepackage{booktabs} 
\usepackage{cite}
\usepackage{algorithmic}
\usepackage{amsmath,amssymb,amsfonts}
\usepackage{textcomp}
\usepackage{xcolor}
\usepackage{xspace}
\usepackage{xurl}
\hypersetup{
  colorlinks,
  linkcolor={blue!70!green},
  citecolor={green!70!blue},
  urlcolor={orange!70!red}
}
\usepackage{caption}
\usepackage{
  color,
  float,
  epsfig,
  wrapfig,
  graphics,
  graphicx,
  subcaption,
  appendix
}

\newcommand{\mypara}[1]{\noindent\textbf{#1.}\xspace}

\newcommand{\MethodName}{GenWatermark\xspace}

\begin{document}

\date{}
\title{\Large \bf Generative Watermarking Against Unauthorized Subject-Driven Image Synthesis}
\author{
Yihan Ma\textsuperscript{1}\ \ \
Zhengyu Zhao\textsuperscript{2}\ \ \
Xinlei He\textsuperscript{1}\ \ \
Zheng Li\textsuperscript{1}\ \ \
Michael Backes\textsuperscript{1}\ \ \
Yang Zhang\textsuperscript{1}
\\
\\
\textsuperscript{1}\textit{CISPA Helmholtz Center for Information Security} \ \ \ 
\textsuperscript{2}\textit{Xi'an Jiaotong University}
}

\maketitle

\begin{abstract}
Large text-to-image models have shown remarkable performance in synthesizing high-quality images.
In particular, the subject-driven model makes it possible to personalize the image synthesis for a specific subject, e.g., a human face or an artistic style, by fine-tuning the generic text-to-image model with a few images from that subject.
Nevertheless, misuse of subject-driven image synthesis may violate the authority of subject owners.
For example, malicious users may use subject-driven synthesis to mimic specific artistic styles or to create fake facial images without authorization.
To protect subject owners against such misuse, recent attempts have commonly relied on adversarial examples to \textit{indiscriminately} disrupt subject-driven image synthesis.
However, this essentially prevents any benign use of subject-driven synthesis based on protected images.

In this paper, we take a different angle and aim at protection without sacrificing the utility of protected images for general synthesis purposes.
Specifically, we propose \MethodName, a novel watermark system based on jointly learning a watermark generator and a detector.
In particular, to help the watermark survive the subject-driven synthesis, we incorporate the synthesis process in learning \MethodName by fine-tuning the detector with synthesized images for a specific subject.
This operation is shown to largely improve the watermark detection accuracy and also ensure the uniqueness of the watermark for each individual subject.
Extensive experiments validate the effectiveness of \MethodName, especially in practical scenarios with unknown models and text prompts (74\% Acc.), as well as partial data watermarking (80\% Acc. for 1/4 watermarking).
We also demonstrate the robustness of \MethodName to two potential countermeasures that substantially degrade the synthesis quality.
\end{abstract}

\begin{figure}[!t]
\centering
\includegraphics[width=1\columnwidth]{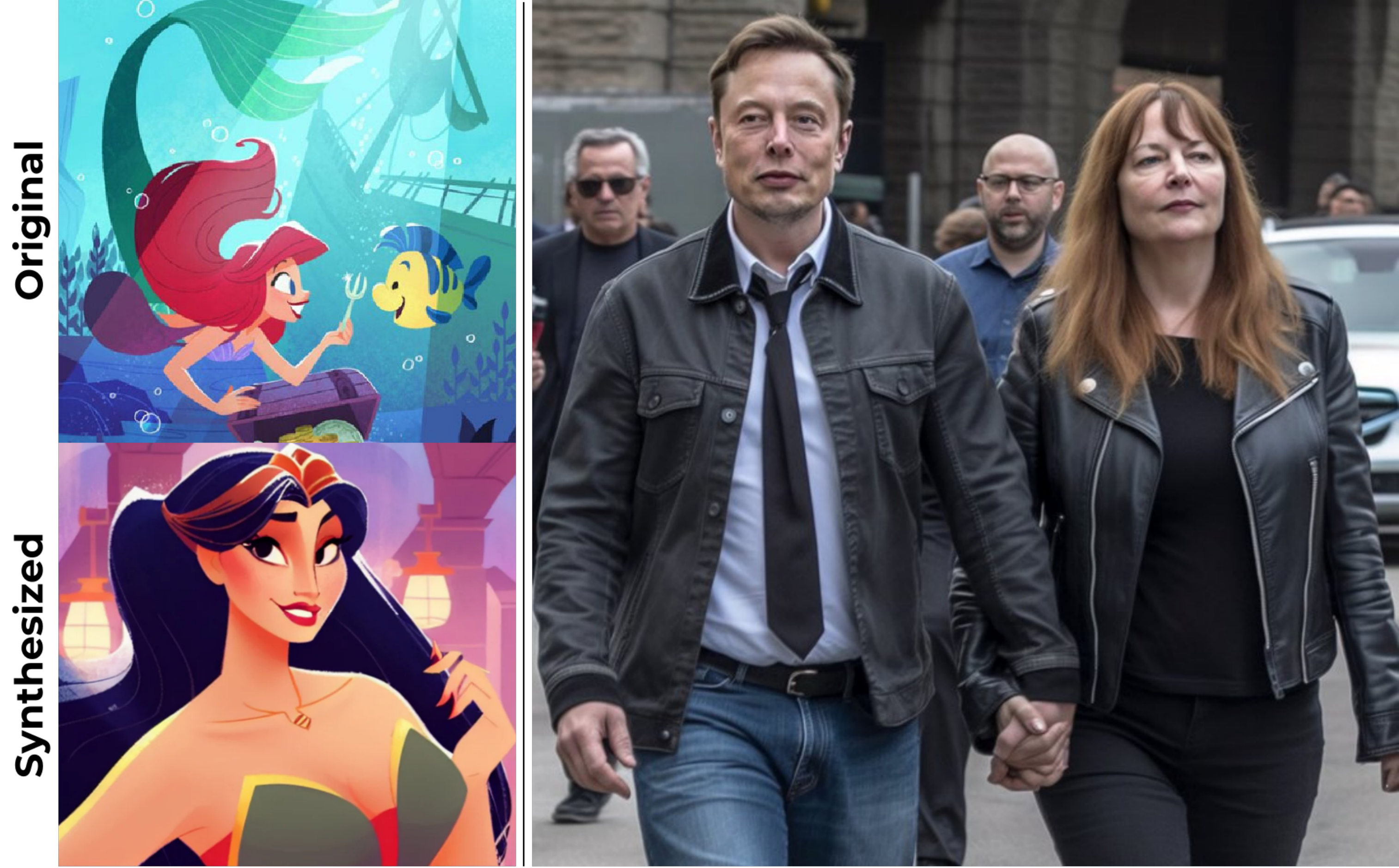} 
\caption{Real-world examples of misuse of subject-driven image synthesis. \textbf{Left}: A piece of original artwork by Hollie Mengert and a synthesized one. \textbf{Right}: A synthesized picture of Elon Musk dating GM CEO Mary Barra.}
\label{figure:elon_musk}
\end{figure} 

\section{Introduction}

Large text-to-image generative models, such as DALL$\cdot$E 2~\cite{RDNCC22}, Latent Diffusion~\cite{RBLEO22}, and Stable Diffusion~\cite{RBLEO22}, have attracted increasing attention~\cite{Midjourney, DD22,BNHVSKAALCKL22,CHSC22,KZLTCDMI22,MGKEHS22}. 
These models take as input a natural language description, referred to as a prompt, and produce high-quality images matching that description.
Text-to-image models have significantly revoluted the AI art industry owing to their substantial generation capacity and ease of use.
For example, an AI-generated image took the first place in the digital category at the Colorado State Fair, beating all (human) artists.\footnote{\url{https://www.nytimes.com/2022/09/02/technology/ai-artificial-intelligence-artists.html}}

One intriguing extension to the generic text-to-image models is the subject-driven synthesis, which makes it possible to ``personalize'' the image synthesis~\cite{GAAPBCC22, RLJPRA23, MHSSWZE22}.
In subject-driven synthesis, the model takes as input a set of images sharing the same subject in addition to the text prompt.
The model learns to synthesize images of that subject in novel contexts, which are specified by the text prompts. 
Popular subject-driven models, such as Textual Inversion~\cite{GAAPBCC22} and DreamBooth~\cite{RLJPRA23} are able to synthesize high-quality images with diverse contents by fine-tuning an off-the-shelf text-to-image model with only a few images.

Subject-driven synthesis serves as an easy-to-use and creative tool for users to synthesize images based on their own needs and potentially benefit from the synthesis~\cite{CIVITAI}.
For example, artists may authorize third-party AI-powered design services to help accelerate the artwork production process, especially in the animation industry.
Individuals can also authorize third-party AI-powered photo editing services to efficiently synthesize images depicting them in varied novel scenes, e.g., tourist attractions.  

However, the images may be leaked and misused to fine-tune a text-to-image model for fabricating images with potentially unwanted content.   
For example, a Reddit user published a DreamBooth model that is fine-tuned based on the artwork from an American artist Hollie Mengert.\footnote{\url{https://www.reddit.com/r/StableDiffusion/comments/yaquby/2d_illustration_styles_are_scarce_on_stable/}}
This has brought huge disputes because anyone can use that model to synthesize artwork in her style \textit{without her authorization}.\footnote{\url{https://waxy.org/2022/11/invasive-diffusion-how-one-unwilling-illustrator-found-herself-turned-into-an-ai-model/}}
In another real-world incident with a person as the subject, a synthesized picture showing Elon Musk is dating GM CEO Mary Barra has been posted on Twitter,\footnote{\url{https://twitter.com/blovereviews/status/1639988583863042050}} as shown in \autoref{figure:elon_musk}.

To protect subject owners against misuse of their images for subject-driven synthesis, concurrent studies~\cite{SCWZHZ23,LWHZXSXMG23,LPNDTT23} rely on adversarial examples, i.e, cloaking the subject owners' images by adding adversarial perturbations.
In this way, the subject-driven synthesis fails because the model instead captures the adversarial image representations rather than the correct ones.
However, such protections are indiscriminate, i.e., they disrupt not only the malicious use of protected images but also their benign use.
Therefore, it cannot address the above problem of unauthorized image synthesis.

\mypara{Our Work} In this paper, instead of completely disrupting subject-driven synthesis on the protected images, we prevent only their unauthorized use, but with little impact on their authorized utility.
Watermarking is intuitively applicable to this task by injecting an invisible watermark into source images such that the owner’s authenticity can be verified by inspecting the watermark in their copies that are potentially manipulated~\cite{B17,TMN20}.
Moreover, the nature of personalized synthesis makes it easy to trace potentially unauthorized (synthesized) images since they share the same subject, i.e., artistic style or human face, with the original images.
However, we find that existing watermarks can hardly survive subject-driven synthesis because the image content (except for the subject) can be freely changed according to the text prompt (see detailed discussions in \autoref{section:watermark}).

Therefore, we propose \MethodName, a novel generative watermarking approach based on a watermark generator and a detector.
Specifically, the generator learns an image watermark in the form of additive perturbations, and the detector tries to distinguish watermarked images from normal images. 

In our evaluation, we conduct extensive experiments to validate the effectiveness of \MethodName in varied scenarios.
We test two well-studied tasks corresponding to two kinds of authority-sensitive subjects, i.e., \textit{artistic style} and \textit{human face}.
In particular, we show that our \MethodName learned on large-scale human face data is effective in both tasks.
Besides, we conduct the evaluation on two representative subject-driven models, i.e., Textual Inversion and DreamBooth, with diverse text prompts. 
In the basic scenario where the subject owner knows the model and prompts the adversary would use, our \MethodName achieves perfect detection accuracy, i.e., above 98\% averaged over two tasks and two models.
Even in the most challenging scenario where neither the model nor prompts are known, \MethodName still guarantees an accuracy of about 74\%, which is much higher than the random chance for binary classification, i.e., 50\%.
Note that the model utility is well maintained in all of the above scenarios.
For example, the Fr\'echet Inception Distance (FID)~\cite{HRUNH17} between the synthesized and input images before and after the watermarking only changes by less than 1\%.
Extensive visualizations of image examples also support this conclusion (see \autoref{figure:example2} as an example). 

We further consider more practical scenarios.
For example, we show that when only half of the input images are watermarked, \MethodName still ensures a detection accuracy of about 90\%. 
Additional results further validate the robustness of our \MethodName to two potential countermeasures.
First, when the adversary deliberately injects random noises into the clean input images, our \MethodName would still classify them as non-watermarked images with high accuracy above 80\%.
Second, when the adversary tries to eliminate the watermarks by image transformations, our \MethodName still achieves a high accuracy above 75\%.
Moreover, we show that the last two countermeasures lead to obvious visual artifacts in both the input and synthesized images.

Our contributions are summarized as follows.
\begin{itemize}
    \item We propose \MethodName, the first watermark approach to protecting images from unauthorized subject-driven image synthesis. Beyond existing methods, \MethodName maintains the utility of the protected images for benign image synthesis.

    \item We demonstrate the effectiveness of \MethodName in two synthesis tasks with two representative subject-driven synthesis models.
    We particularly consider realistic scenarios with unknown target models and prompts, as well as with only partial watermarked input images.
    
    \item We demonstrate the robustness of \MethodName against two potential countermeasures, which are based on adding noise to mislead the detector or using image transformations to disrupt our watermarks.
    
\end{itemize}

\section{Related Work}

\subsection{Protection Against Malicious Subject-Driven Image Synthesis}
\label{section:related}

Although subject-driven models make it possible for users to synthesize images easily based on their own needs, the potential malicious use of such tools may cause severe threats.
As discussed, adversaries may use subject-driven synthesis to mimic the artistic style of specific artists or to generate fake images containing specific faces.
To protect users against such malicious use, existing studies~\cite{SCWZHZ23,LWHZXSXMG23,LPNDTT23} have commonly relied on adversarial examples~\cite{SZSBEGF14,GSS15}.
The basic idea is to add carefully crafted perturbations to the users' images such that the subject-driven models cannot learn the correct image representations anymore during model fine-tuning.
These three studies are different in their technical details.
Specifically, both AdvDM~\cite{LWHZXSXMG23} and Anti-DreamBooth~\cite{LPNDTT23} assume access to the diffusion process of the target model, and (approximate) gradients can be computed for perturbation optimization.
Differently, Glaze~\cite{SCWZHZ23} assumes no access to the diffusion process but only a generic image feature extractor.
In this case, the perturbations are optimized to cause disruptions in the latent space of that feature extractor.
Glaze is specifically designed to protect artists by concentrating the perturbations on artistic style-specific features that are isolated based on a style transfer model.

A common limitation of the above approaches is that they do not discriminate the malicious use of subject-driven synthesis from its benign use.
Instead, they completely disable the subject-driven synthesis on the protected images.
Our work is essentially different because we focus on preventing only malicious use of subject-driven synthesis while maintaining the possibility for potential benign use.
In addition, our \MethodName is the first approach to protecting images against subject-driven synthesis for both tasks of human face identity and artistic style, and it remains highly effective even in cross-task settings.
Technically, it requires no knowledge about the internal details (e.g., diffusion process and gradients) of the target model but only use the output images.

\subsection{Image Watermarking}
\label{section:watermark}
Traditional image watermarking mainly relies on steganography~\cite{B142}, achieved by Fourier transformation~\cite{CMB02,CFF05}, JPEG compression~\cite{outguess}, or least significant bits modification~\cite{PFB10,HFD14}.
Among them, Fourier transformation and JPEG compression are known to cause visible visual artifacts, making them not inappropriate in general.
The least significant bits modification method hides the watermark in the last bit of each pixel, which is invisible but not robust to even simple noise.
These traditional watermarks have been shown to be ineffective in the context of generative synthesis~\cite{YSAF21}.
Therefore, deep learning-based image steganography methods based on an encoder and a decoder were proposed~\cite{B17,HD17,VCF18,ZKJF18,TMN20,YSAF21}. 

These approaches were originally designed against generative adversarial networks (GANs)-based image synthesis, and we find it does not work well in our context with the subject-driven model.
In particular, our exploratory results demonstrate that the state-of-the-art approach from~\cite{B17,HD17,VCF18,ZKJF18,TMN20,YSAF21} only achieves a bit-wise accuracy around the random guess level, i.e., 50\%.
We hypothesize that there are two main reasons.
First, the training of the subject-driven image synthesis only requires a small number of images (e.g., 30 in our work), which is not enough to train a useful image steganography encoder and decoder.
Second, with the semantic guidance of the text prompt, subject-driven image synthesis leads to more diverse changes of image content than the GAN tasks considered in~\cite{YSAF21}, such as deepfakes and image-to-image translation.
One recent preprint~\cite{FCJDF23} also explores watermarking techniques for text-to-image diffusion models.
However, ~\cite{FCJDF23} does not consider a subject-driven task, so their authority is defined with respect to the model instead of a subject. 

\section{Preliminary}

\subsection{Text-to-Image Synthesis}

Since first proposed in 2015~\cite{MPBS16}, text-to-image models have been widely explored and used for image synthesis~\cite{RDNCC22, DYHZZYLZSYT21, JYXCPPLSLD21, LZZHHLG19, RMC16, ZXL17, RBLEO22,XZHZGHH18, Dayma_DALL·E_Mini_2021,TTWJBX22, ZPCY19}.
In text-to-image synthesis, natural language descriptions, namely \textit{prompts}, are used as the input to generate synthetic images.
Normally, text-to-image models contain two parts: a text embedding module, e.g., CLIP's~\cite{RKHRGASAMCKS21} text encoder or BERT~\cite{DCLT19}, that turns natural language into corresponding semantic features, and a conditional image generator that produces synthetic images based on the semantic features.
Conventional work leverages generative adversarial networks (GANs) or variational autoencoders (VAEs) as the image generator~\cite{RDNCC22, DYHZZYLZSYT21, RMC16, Dayma_DALL·E_Mini_2021, TTWJBX22, ZPCY19}, but recent work has been shifting to diffusion models and shows great advantages in producing high-quality images with rich details~\cite{RDNCC22, RPGGVRCS21, RBLEO22, SCSLWDGAMLSHFN22}.

\mypara{Diffusion Models}
The core of a diffusion model is a stochastic differential process called the \textit{diffusion process}, which is composed of 2 phases: the forward diffusion process and the backward denoising process~\cite{SWMG15}.
Given a distribution of real images $q(\mathbf{x})$, the forward diffusion process employs a Markov chain of diffusion steps to slowly add random noise $\epsilon$ to 
a data sample $\mathbf{x}_{0} \sim q(\mathbf{x})$ in $\mathbf{T}$ steps, producing a sequence of noised samples $\mathbf{x}_{1}, \ldots, \mathbf{x}_{T}$:
\begin{equation}
    \mathbf{x}_{T}=\sqrt{1-\beta_{t}} \mathbf{x}_{t-1}+\sqrt{\beta_{t}} \epsilon.
\end{equation}
Here, the step sizes are controlled by a variance schedule $\left\{\beta_{t} \in(0,1)\right\}_{t=1}^\mathbf{T}$, and the noise $\epsilon \sim \mathcal{N}(0, \mathbf{I})$ is sampled from the Gaussian distribution.
Note that the data sample becomes more unrecognizable as with a larger step $\mathbf{T}$.
As a result, when $\mathrm{T} \to \infty$, $\mathbf{x}_{T}$ will follow the isotropic Gaussian distribution.

The backward denoising process is the reverse of the forward diffusion process, which is to generate $\mathbf{x}_{T}$ given $\mathbf{x}_{T+1}$.
With $\mathbf{T}$ being sufficiently large, we are able to create a true sample from $q(\mathbf{x})$ by a Gaussian noise input $\mathbf{x}_T \sim \mathcal{N}(0, \mathbf{I})$.
The key to this process is to train a neural network $\epsilon_\theta (\mathbf{x}_{t+1}, t)$ that can estimate the injected noise $\epsilon$ in step $t$.
Specifically, this denoising neural network is trained to minimize the L2 distance between the real noise and the estimated noise using the following loss function:
\begin{equation}
    \mathcal{L}(\theta)=\mathbb{E}_{t, \mathbf{x}_{0, \varepsilon} \sim \mathcal{N}(0,1)}\left[\left\|\varepsilon-\varepsilon_{\theta}\left(\mathbf{x}_{t+1}, t\right)\right\|_{2}^{2}\right],
\end{equation}
where $t$ is uniformly sampled from all time steps $\{1,\ldots,\mathbf{T}\}$.
So far, there are two representative diffusion-based text-to-image models: Latent Diffusion and Stable Diffusion.
\begin{itemize}
    \item \textbf{Latent Diffusion} runs the diffusion process in the latent space instead of the pixel level, which leads to less training cost and faster inference speed while maintaining high-quality generation.
    \item \textbf{Stable Diffusion} is a variant of Latent Diffusion and they share a similar architecture. The major difference is that the text encoder in Stable Diffusion is the CLIP's text encoder~\cite{RKHRGASAMCKS21} instead of BERT~\cite{DCLT19}.
\end{itemize}

\subsection{Subject-Driven Image Synthesis}

Subject-driven image synthesis leverages the basic text-to-image models and takes additional images as the input.
Given a few images of the same subject, the subject-driven image synthesis can learn the subject in the input samples, and synthesize novel renditions of the learned subject in different contexts, using different prompts to guide the generation direction.
In real-world applications, the learned subject can either be the face of a person, or the artistic style of an artist, as shown in \autoref{figure:elon_musk}.

The workflow of the subject-driven image synthesis is normally as follows.
First, subject-driven image synthesis learns the distribution of the subject shared in a set of images.
The learned distribution can be represented by a pseudo-word, referred to as ``$[\mathbf{V}]$''.
Once learned, a new prompt containing ``$[\mathbf{V}]$'' can be used as input to synthesize new images depicting the same subject with different contexts. 
So far, there are two popular subject-driven image synthesis models: Textual Inversion~\cite{GAAPBCC22} and DreamBooth~\cite{RLJPRA23}.
Their detailed information is described as follows.

\mypara{Textual Inversion} 
Textual Inversion learns the embedding of the subject ``$[\mathbf{V}]$'' while freezing the parameters of the text-to-image model, here, Latent Diffusion.
The training takes input pairs of ``$[\mathbf{V}]$'' and images.
During synthesis, if we want to synthesize a painting of a daisy in the same style of an artist $[\mathbf{V}]$, the new prompt could be something like \textit{``A painting of daisy in the style of $[\mathbf{V}]$''}.

\mypara{DreamBooth}
Unlike textual inversion, DreamBooth is based on fine-tuning model parameters and uses Stable Diffusion as the backbone.
Specifically, the prompt is designed to contain an additional term denoting the category of the subject.
In this case, the above prompt example becomes \textit{``A painting of a daisy in the style of $[\mathbf{V}]$ painting''}.

\section{Methodology}
\label{section:methodology}

In this section, we introduce the methodology of our new watermark method \MethodName.
Before introducing the technical details \MethodName, we first specify the threat model and discuss the design intuition, where we also identify two potential challenges in designing an effective watermarking approach in the context of subject-driven synthesis.
\begin{figure}
    \centering
    \includegraphics[width=\columnwidth]{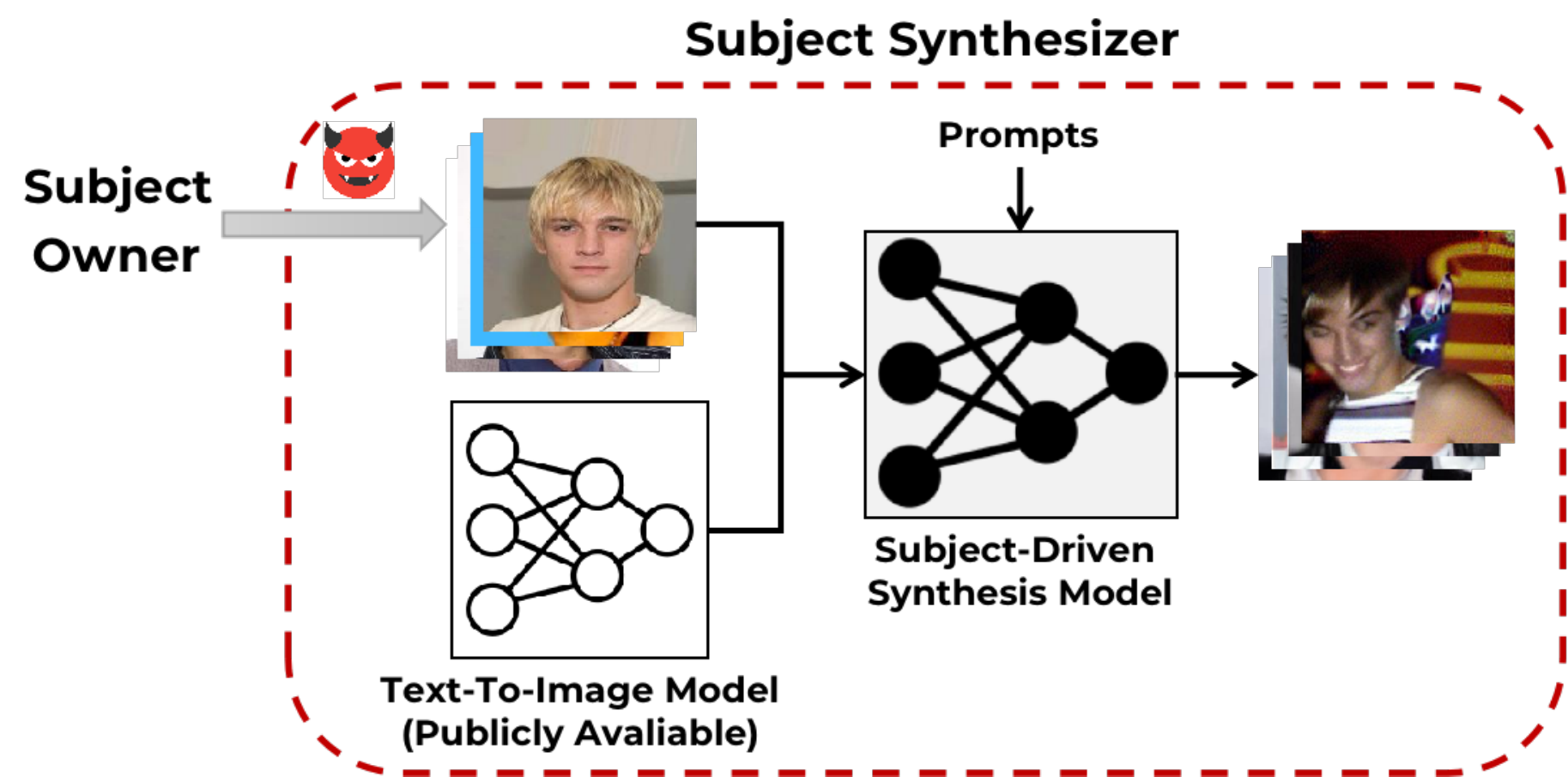}
    \caption{High-level overview of subject-driven image synthesis attack scenario.}
    \label{figure:attackscenario}
\end{figure}

\subsection{Threat Model}
\label{section:threatmodel}

In our threat model, there are two parties involved: the subject owners and the subject synthesizers.
Subject owners use the generator of \MethodName to obtain watermarks and inject them into their images before authorizing specific subject synthesizers to use those images.
Subject synthesizers synthesize images targeting the protected subject by fine-tuning a subject-driven synthesis on those watermarked images.
Adversaries refer to the subject synthesizers who do not obtain authorization from the owners.
\autoref{figure:attackscenario} provides a high-level overview of our attack scenario.

\mypara{Subject Owners} 
The goal of the subject owner is to benefit from authorizing specific third parties to use their images for subject-driven synthesis.
For example, artists may authorize AI-powered design services to assist their artwork production.
The subject owners use our system to generate the watermark and inject it into their images.
Then, they can track the potential unauthorized use by detecting if the watermark appears in synthesized images.

We assume the subject owner has full access to our watermark generator and detector.
They can further improve the watermark detector by fine-tuning it on additional images that are synthesized using a popular subject-driven synthesis service, such as Textual Inversion and DreamBooth.
For the knowledge of the subject-driven model and text prompts, we consider 4 scenarios with gradually relaxed assumptions: 1) Both are known; 2) Only the model is known; 3) Only the prompts are known; 4) Neither is known.

\mypara{Subject Synthesizers} 
The goal of the subject synthesizer is to train a subject-driven model that synthesizes high-quality images of the target subject.
A benign subject synthesizer obtains authorization from the subject owner but a malicious subject synthesizer achieves the goal without authorization.
For example, the malicious subject synthesizer can mimic the style of an artist based on the leaked artwork from third-party AI-powered design services.
We assume the subject synthesizer has access to a publicly-available subject-driven model and also protected images from the target subject.
The subject synthesizer is also assumed to have sufficient computational resources to fine-tune the model.

\begin{figure*}[!t]
\centering
\includegraphics[width=0.9\textwidth]{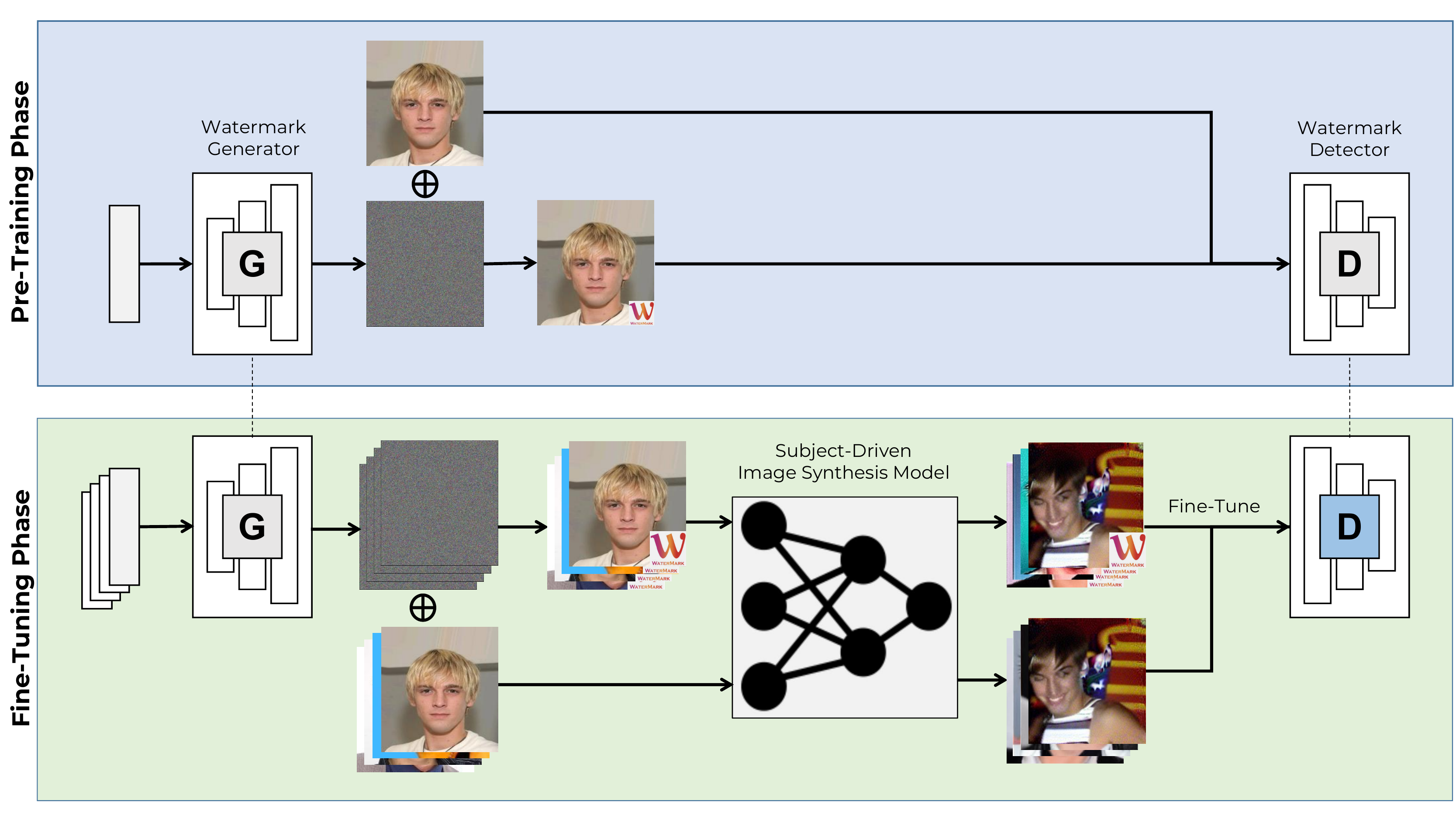} 
\caption{Overview of \MethodName. The pipeline of \MethodName can be divided into 2 phases. In the first phase, we train a watermark generator and detector using a large-scale dataset. The trained watermark generator is then used to produce watermarked images for the subject we want to protect. In the second phase, each subject is allowed to ``personalize'' the pre-trained detector by fine-tuning it on images synthesized from their own images.}
\label{figure:overview}
\end{figure*} 

\subsection{Design Intuition}

In general, a successful watermark requires the \textit{robustness} and \textit{invisibility}~\cite{YSAF21}.
In our context of subject-driven synthesis, robustness means the watermark should be robust to substantial changes in the image content caused by the subject-driven text-to-image generation, and invisibility means the watermark should have little impact on the quality of both the watermarked (input) images and synthesized (output) images.

More specifically, we identify two potential challenges in achieving an effective and practical watermark system for subject-driven image synthesis.
The first challenge is to incorporate the image synthesis process into the optimization of our watermark system.
Existing protections based on adversarial perturbations~\cite{LWHZXSXMG23,LPNDTT23} incorporate the internal (diffusion) process of the target model and optimize their perturbations based on (approximate) gradients.
This is computationally inefficient and requires knowledge of the diffusion process.
However, it makes more practical sense to have a protection method that exhibits high efficiency and ease of optimization since the subject owner, e.g., an artist, does not have large computational resources and rich knowledge of computer science.
Another specific reason for incorporating the image synthesis process is that the image synthesis process captures the unique information for each individual subject, and we want to use such information to achieve a subject-personalized watermark design.

The second challenge is to maintain the watermark's effectiveness in realistic scenarios with unknown settings.
Specifically, the watermark system should generalize well to different target models and text prompts because the subject owner has no control of the subject-driven model and text prompts that will be used by the malicious subject synthesizer for their unauthorized image synthesis.
In addition, the watermark system should generalize well to different data distributions.
For example, using only a few pieces of artwork from a specific artist may not be enough to train a powerful watermark system for that artist.

\subsection{Our \MethodName}
We propose \MethodName, a generative watermarking approach against unauthorized subject-driven synthesis.
To achieve \textit{robustness}, \MethodName follows a generative framework consisting of a watermark generator and a detector.
The generator and the detector are jointly trained such that the generator can generate strong watermarks and the detector can detect the watermark left in synthesized images.
To achieve \textit{invisibility}, we follow related work~\cite{SCWZHZ23, CGFDDTG21, LSF21, RGPA21} to constrain the perturbation budget measured by some perceptual metrics.

In particular, we aim to address the above two potential challenges.
For the first challenge, we incorporate the process of subject-driven synthesis into our \MethodName by allowing each subject to fine-tune the detector on their own.
The fine-tuning dataset consists of synthesized images that are obtained from the subject-driven synthesis using both clean and watermarked images with diverse text prompts.
For the second challenge, the fact that \MethodName is not optimized based on the internal details of the subject-driven synthesis makes it compatible with unknown target models.
In addition, we develop \MethodName with diverse prompts and gather images from multiple subjects or even from other domains with sufficient public data, such as the face image domain.

As shown in \autoref{figure:overview}, the whole learning pipeline can be divided into 2 phases: pre-training and fine-tuning.

\mypara{Phase 1: Pre-training the generator and detector}
In this phase, we jointly train a watermark generator $\mathbf{G}$ and a watermark detector $\mathbf{D}$.
For initialization, we randomly choose a latent code $\mathbf{z}$~\cite{GPMXWOCB14, KLA19, CCKHKC18} as the input to generate a watermark $\mathbf{w}$ with the same size as the input image.
Here we adopt the well-known GAN generator~\cite{GPMXWOCB14} as our watermark generator.

During training, this watermark $\mathbf{w}$ is added to each original (clean) image $\mathbf{x}$ to obtain the corresponding watermarked image $\mathbf{x}_w$: 
\begin{equation}
    \mathbf{x}_w = \mathbf{x} \oplus \mathbf{w}
\end{equation}
The Learned perceptual image patch similarity (LPIPS)~\cite{ZIESW18} is engaged to constrain the invisibility of the watermark when training the generator $\mathbf{G}$.
LPIPS measures the distance between image patches represented in the latent space of a conventional neural network (CNN)~\cite{RGPA21, LSF21, CGFDDTG21}.
LPIPS is known to align better with human visual perception than other perceptual metrics, such as the naive $L_p$ distance and SSIM~\cite{ KGB16,SCFF16, CW17,JG18}. 
The optimization of the generator is as follows:
\begin{equation}
\label{equ:p}
    \mathcal{L}_G = \max(LPIPS(\mathbf{x}, \mathbf{x}_w)-p, 0),
\end{equation}
where $p$ controls that desired level of invisibility, and a smaller $p$ represents lower visibility.

The detector $\mathbf{D}$ is trained to distinguish between clean and watermarked images.
Specifically, we adopt a convolutional neural network (CNN) for the detector and the binary cross-entropy (BCE) loss for optimization.
\begin{equation}
    \mathcal{L}_D = -(1-y) \log (1-\hat{y})-y \log (\hat{y}),
\end{equation}
Where $y$ represents the ground truth and $\hat{y}$ means the predicted value.
Then, the overall loss function can be formulated as:
\begin{equation}
\label{equ:alpha}
    \mathcal{L} = \alpha \cdot \mathcal{L}_G + \mathcal{L}_D,
\end{equation}
where $\alpha$ is a hyperparameter to balance the two different objectives.
Finally, we use the pre-trained generator $\mathbf{G}$ to watermark images for the subject we want to protect.

\mypara{Phase 2: Fine-tuning the detector}
In this phase, we further improve the performance of the detector $\mathbf{D}$ by allowing each individual subject to ``personalize'' the detector on their own images. 
First, we use a clean image set $\mathbf{X}$ and its corresponding watermarked set $\mathbf{X}_w$ to train two subject-driven models, i.e., a clean model $\mathcal{M}$ and a watermarked model $\mathcal{M}_w$.
Then, we use these two models to synthesize images with multiple prompts, resulting in two corresponding synthesized image sets, denoted as $\mathbf{S}$ and $\mathbf{S}_{w}$.
Finally, we use $\mathbf{S}$ and $\mathbf{S}_{w}$ to fine-tune the detector.
Note that the generator remains unchanged.

In this way, protecting each individual subject becomes a downstream task that can be solved efficiently.
Specifically, this fine-tuning has two main advantages.
First, it improves the watermark detection performance by incorporating the process of subject-driven synthesis. See \autoref{section:ablation} for experimental results. 
Second, it isolates the protections of different subjects, making sure that the authority of images is hard to be claimed by anyone except the real owner.
See more relevant discussions in \autoref{section:counter}.

\section{Experiments}

In this section, we evaluate the performance of our proposed \MethodName against malicious subject-driven image synthesis.
First, we present detailed information about experimental settings.
Then, we evaluate the performance of \MethodName from two perspectives: the watermark detection accuracy and the image synthesis quality.
Furthermore, we conduct comprehensive ablation studies to analyze the effects of the attack assumptions and main hyperparameters.

\subsection{Experiment Setup}
\label{section:experimentsetup}

All our experiments were run on one NVIDIA A100
GPU.
Specifically, the pre-training phase needs to run only one time and takes about 6 hours.
In the fine-tuning phase, the training of the subject-driven model is run independently for each subject, and it takes about 30 minutes for Textual Inversion and 2 hours for DreamBooth.
The inference time is about 10 minutes.
Then, the fine-tuning of our watermark detector for each subject takes about 1 hour.
In sum, for each subject owner, the fine-tuning phase takes less than 4 hours.

\mypara{Tasks and Datasets}
We consider two well-studied tasks corresponding to two kinds of subjects: \textit{artistic style} and \textit{human face}~\cite{SCWZHZ23, LPNDTT23, SKLIM23}.
In the human face task, we use the large-scale dataset CelebA~\cite{LLWT15}, which contains 202,599 face images of 10,177 celebrities.
In this case, the subject-driven model is used to learn the unique face of each celebrity [$\mathbf{V}$] and synthesize new images of the celebrity in different scenes.
    
In the artistic style task, we use WikiArt~\cite{SE15}, which contains 52,757 paintings from 195 different artists in 27 genres.
Each painting is associated with a caption that describes the content.
In this case, the subject-driven model is used to learn the unique artistic style of each artist [$\mathbf{V}$] and synthesize new images in the same style with different contents.

\mypara{Prompt Design}
We design text prompts involving diverse content.
In the human face task, we randomly select 30 prompts from Lexica, a popular search engine that contains millions of AI-synthesized images and their corresponding prompts.\footnote{\url{https://lexica.art/}}
We refine each prompt by removing irrelevant information but only keeping a target context to form our final prompt following \textit{``A photo of [$\mathbf{V}$] {\color{blue}{target context}}''}.
 For example, for an original prompt \textit{``smiling softly, 8k, irakli nadar, hyperrealism, hyperdetailed, ultra realistic''}, our refined prompt is \textit{``A photo of [$\mathbf{V}$] {\color{blue}{smiling softly}}''}.
Since DreamBooth requires an additional term denoting the category of the subject, the prompt becomes \textit{``A photo of [$\mathbf{V}$] \textbf{face}  {\color{blue}{smiling softly}}''}

In the artistic style task, we randomly select 30 prompts (image descriptions) from WikiArt.
Then, we refine the prompt to follow \textit{``A painting of {\color{magenta}{target description}} in the style of [$\mathbf{V}$]''.}
For example, the prompt can be \textit{``A painting of {\color{magenta}{the valley of the river slavyanka}} in the style of [$\mathbf{V}$]''}.
For DreamBooth, the prompt becomes \textit{``A painting of {\color{magenta}{the valley of the river slavyanka}} in the style of [$\mathbf{V}$] \textbf{painting}''}.
\autoref{table:all_prompts} in \autoref{section:appendixa} lists all 30 prompts used in the human face task and 30 prompts used in the artistic style task.

\mypara{Synthesis Setup}
In the human face task, we construct the training set by randomly selecting 4 celebrities that have more than 30 images.
The IDs of these celebrities are 14, 15, 17, and 21 in CelebA.
In the artistic style task, we select a total of 27 representative artists corresponding to all 27 different genres.
\autoref{table:artists} in \autoref{section:appendixa} lists the names of the selected artists with their corresponding genres.
For each of these selected subjects, we randomly sample 30 images to train their (clean or watermarked) subject-driven models for 6500 steps following the previous work~\cite{GAAPBCC22}.
In total, we get 8 (4 clean and 4 watermarked) models for the human face task and 54 (27 clean and 27 watermarked) models for the artistic style task.
For each of these models, we will synthesize images for each of the 30 prompts in our following experiments.  
The resolution of the synthesized image is 256x256.

\mypara{Watermark Setup}
As mentioned in \autoref{section:methodology}, the training pipeline of our \MethodName consists of two phases: pre-training and fine-tuning.
We adopt the generator from the vanilla GAN~\cite{GPMXWOCB14} as the watermark generator and ResNet34~\cite{HZRS16} as the detector.
In the pre-training phase, we jointly train the watermark generator and detector with 200,000 images from CelebA since a large number of images are required for an effective \MethodName.
In our experiments, we show that this \MethodName trained on CelebA is also effective in watermarking WikiArt paintings.
All the training images are resized to 256x256.
We choose the invisibility level $p=0.05$ and the balancing factor $\alpha=1$.
The effects of these hyperparameters on the performance of \MethodName will be discussed in \autoref{section:ablation}.
In the fine-tuning phase, for each subject, we fine-tune the detector with 1,000 synthesized images (i.e., 40 different images for 25 prompts) from the clean model $\mathcal{M}$ and also 1,000 synthesized images from the watermarked model $\mathcal{M}_w$.

\mypara{Attack Scenarios}
As discussed in the threat model (\autoref{section:threatmodel}), we consider 4 scenarios with gradually relaxed assumptions on the knowledge the subject owner has about the model and prompts the adversary (subject synthesizer) would use.
Their settings are specified as follows.

In \textbf{Scenario 1}, both the model and the prompts are known.
Here we synthesize another 10 images using the same model for each of the same 25 prompts as in our detector fine-tuning.

In \textbf{Scenario 2}, only the model is known but the prompts are unknown.
Therefore, different from Scenario 1, we synthesize another 50 images for each of the rest 5 prompts (out of 30).

In \textbf{Scenario 3}, only the prompts are known but the model is unknown.
Therefore, different from Scenario 1, we use Textual Inversion (DreamBooth) for the adversary but DreamBooth (Textual Inversion) for our detector fine-tuning.

In \textbf{Scenario 4}, neither the model nor the prompts are known.
Therefore, both the model and the prompts differ following Scenario 2 and Scenario 3.

\mypara{Evaluation Metrics}
We evaluate the performance of our \MethodName from two perspectives.
First, we want \MethodName designed for a specific subject to have high detection accuracy on that subject.
In addition, it should achieve uniqueness, i.e., not falsely detecting the watermarked images generated for other subjects.
Second, we want the subject-driven model that is trained on watermarked images to maintain its synthesis quality.
Third, we want the watermark detection to be unique to specific owners, i.e., the watermark detector designed for one subject should not falsely detect the watermark generated for another subject.

\begin{itemize}
    \item \textbf{Watermark Detection Accuracy} is defined as the fraction of correctly classified synthetic images among all the test synthetic images. \textbf{Uniqueness} is measured by the accuracy of a subject's detector in differentiating their own watermarked images from other subjects' watermarked images.
    \item \textbf{Image Synthesis Quality} is measured based on the Fr\'echet Inception Distance (FID)~\cite{HRUNH17} calculated between synthesized images and input images before and after the watermarking.
    If the FID scores before and after watermarking are similar, the watermark is considered to have little impact on the image synthesis quality.
\end{itemize}

\subsection{Results on Watermark Detection Accuracy}
\label{section:result_acc}

\begin{figure}
    \centering
    \includegraphics[width=\columnwidth]{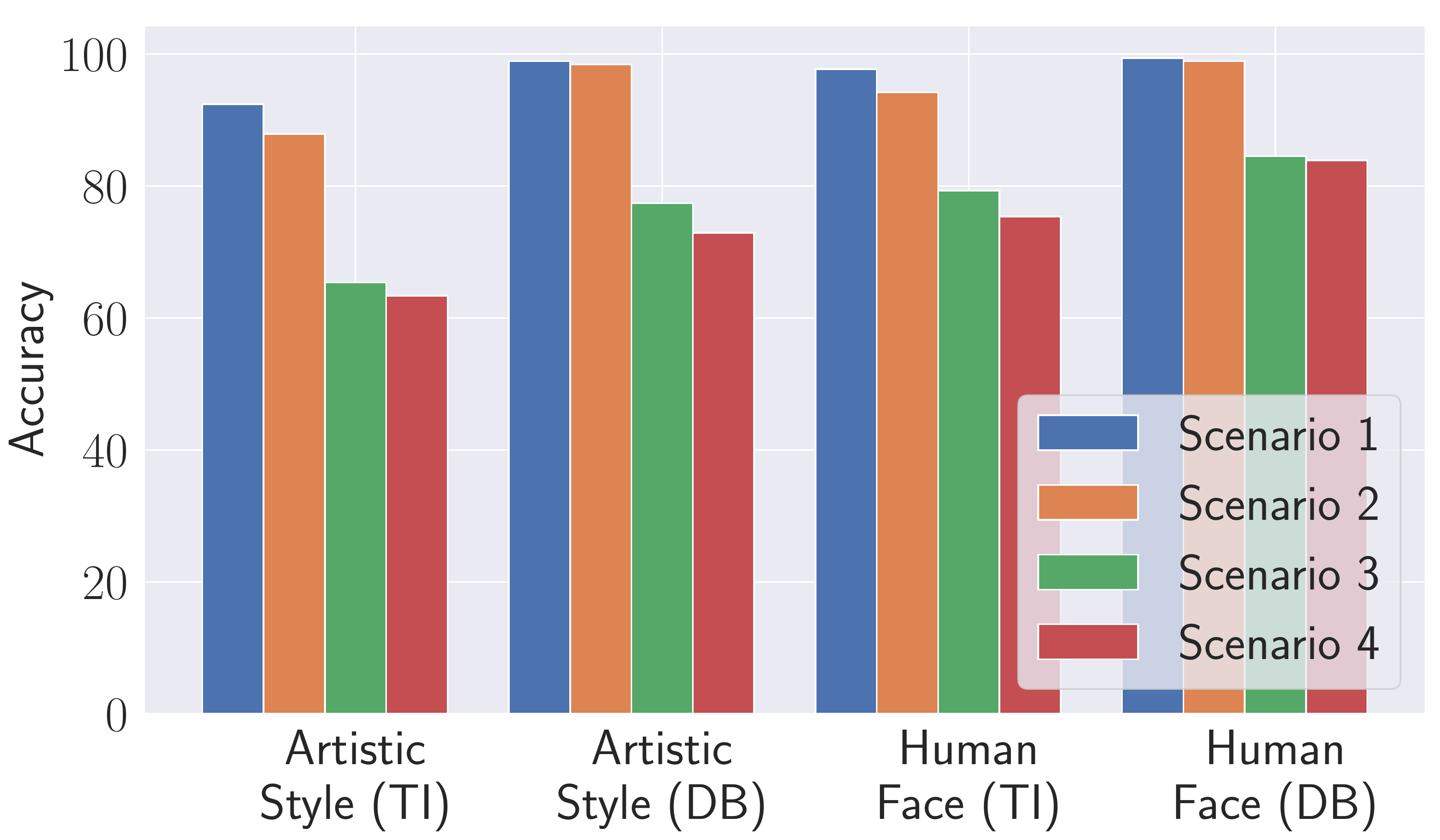}
    \caption{Watermark detection accuracy (\%) on Textual Inversion and DreamBooth in the artistic style and human face tasks considering four attack scenarios.}
    \label{figure:meanaccuracy}
\end{figure}

\mypara{Overall Results}
\autoref{figure:meanaccuracy} presents the overall detection accuracy on two target models for both the human face and artistic style tasks in four different scenarios, as introduced in \autoref{section:threatmodel} and set up in \autoref{section:experimentsetup}.
As can be seen, in Scenario 1, where the model and prompts the adversary would use are both known to the subject owner, our \MethodName achieves perfect results with an accuracy of above 98\% on average.
Even in Scenario 4, the most challenging one with unknown models and prompts, \MethodName still maintains an accuracy of about 74\%, much higher than the random chance for binary classification, i.e., 50\%.
As expected, in Scenario 2 or Scenario 3, where either the model or prompts are known, the performance is in between the two extreme scenarios.
When comparing Scenario 2 and Scenario 3, we can see that knowing the model is much more helpful than knowing the prompts. In other words, transferring between models is more difficult than between prompts.
Overall, results for Scenarios 2-4 suggest the good transferability of \MethodName to unknown models and prompts. 

\begin{figure}
    \centering
    \includegraphics[width=\columnwidth]{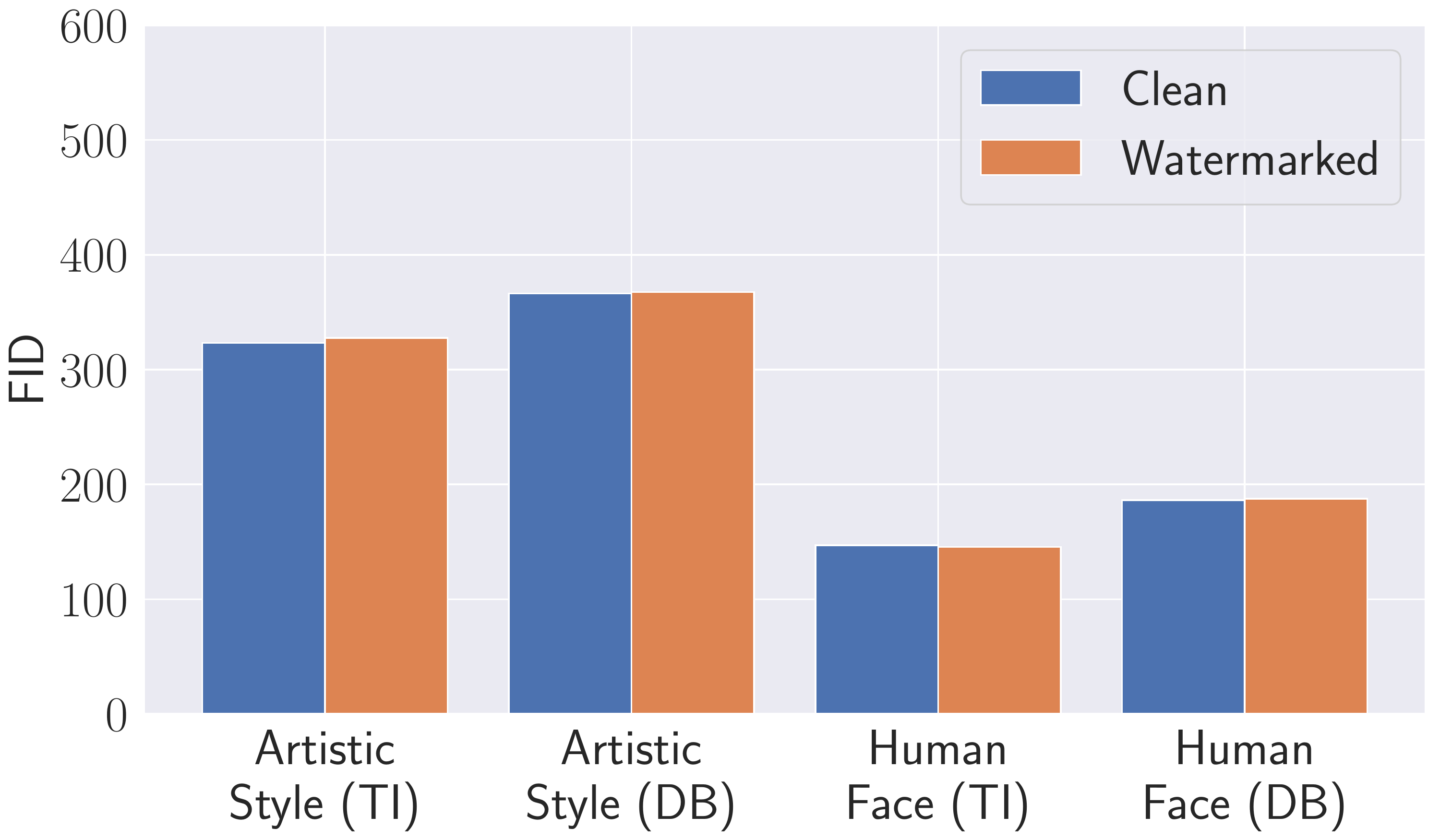}
    \caption{Image synthesis quality measured by the FID score for synthesis with clean inputs and with watermarked inputs. The FID is calculated between the (clean or watermarked) inputs and synthesized outputs.}
    \label{figure:fid}
\end{figure}
 
\begin{figure}[!t]
\centering
\includegraphics[width=\columnwidth]{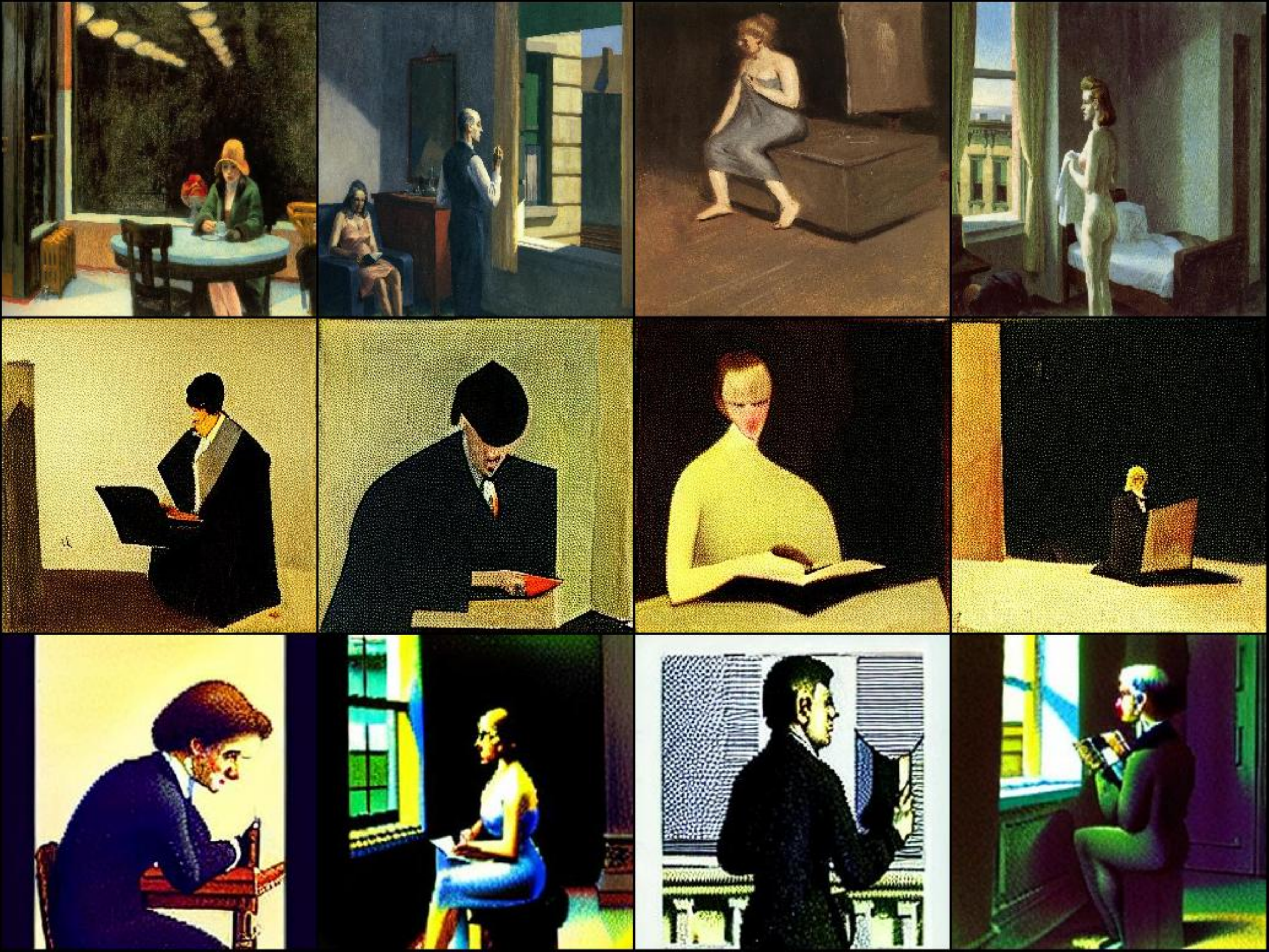} 
\caption{Images synthesized based on the original artwork from Edward Hopper (top) by Textual Inversion (middle) and DreamBooth (bottom). The prompt content is \textit{``A painting of the reader in the style of [V]''}.}
\label{figure:example1}
\end{figure}

\mypara{Different Tasks}
For the two tasks, we can see that the results for the artistic style are consistently worse than for the human face. 
This could be because we use the CelebA face data for pre-training in both tasks.
Moreover, since the artistic style synthesis leads to more severe changes in image content, i.e., the watermark is disrupted more in the artistic style task.
Consistently, from \autoref{figure:fid}, we observe that FID between the input images and synthesized images for the artistic style is about 2-3 times higher than that for the human face.

\mypara{Different Target Models}
For the two target models, we can see that the results for DreamBooth and Textual Inversion are quite different, especially in the artistic style task.
To shed more light on this difference, we visualize examples of synthesized images from these two models in \autoref{figure:example1}.
As can be seen, DreamBooth images inherit fine-grained content from the input images, such as the tables and windows.
The original work of DreamBooth~\cite{RLJPRA23} also shows its power in learning concepts.
Differently, Textual Inversion images mainly capture the holistic style, and their content solely depends on the text prompts with little influence by the fine-grained content in the input images.
This difference also supports our above finding that transferring between models, as shown by the results for Scenario 3 and Scenario 4, is relatively difficult.

\begin{figure}[!t]
    \centering   
    \includegraphics[width=\columnwidth]{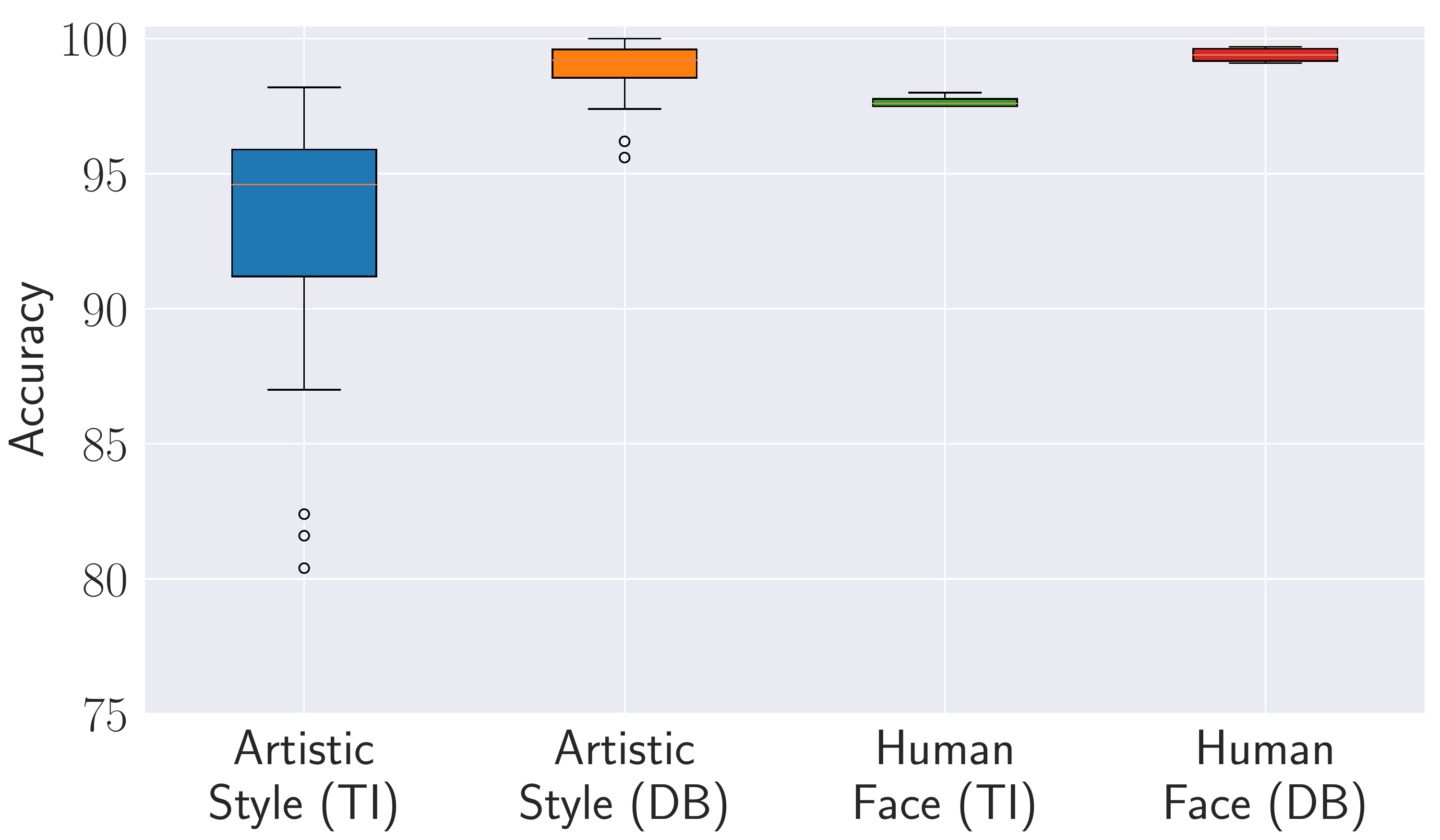}
    \caption{Watermark detection accuracy (\%) for Scenario 1.
     Results for scenarios 2-4 can be found in \autoref{figure:s2s3s4}.}
    \label{figure:s1_artist}
\end{figure}

\mypara{Different Subjects}
All the above results are averaged on all 27 artists in the artistic style task and 4 celebrities in the human face task.
Since each synthesis model is driven independently by an individual subject, we are also interested in the performance distribution of \MethodName over different subjects.
As can be seen in \autoref{figure:s1_artist}, the human face task yields very stable results while the artistic style task yields result with larger variances.
This is understandable because artistic styles selected from different genres are very different, but the human face images are mostly with a fixed front view. 
In addition, DreamBooth yields more stable results than Textual Inversion.
This might be due to the fact that, as discussed above, DreamBooth also learns additional, finer-grained information beyond the holistic style.
In the future, one possible way to further improve the detection accuracy for each subject is to adapt the training strategy, e.g., the number of training images, learning rates, and training epochs, to specific characteristics of different subjects.

\begin{figure}[!t]
\centering
\includegraphics[width=\columnwidth]{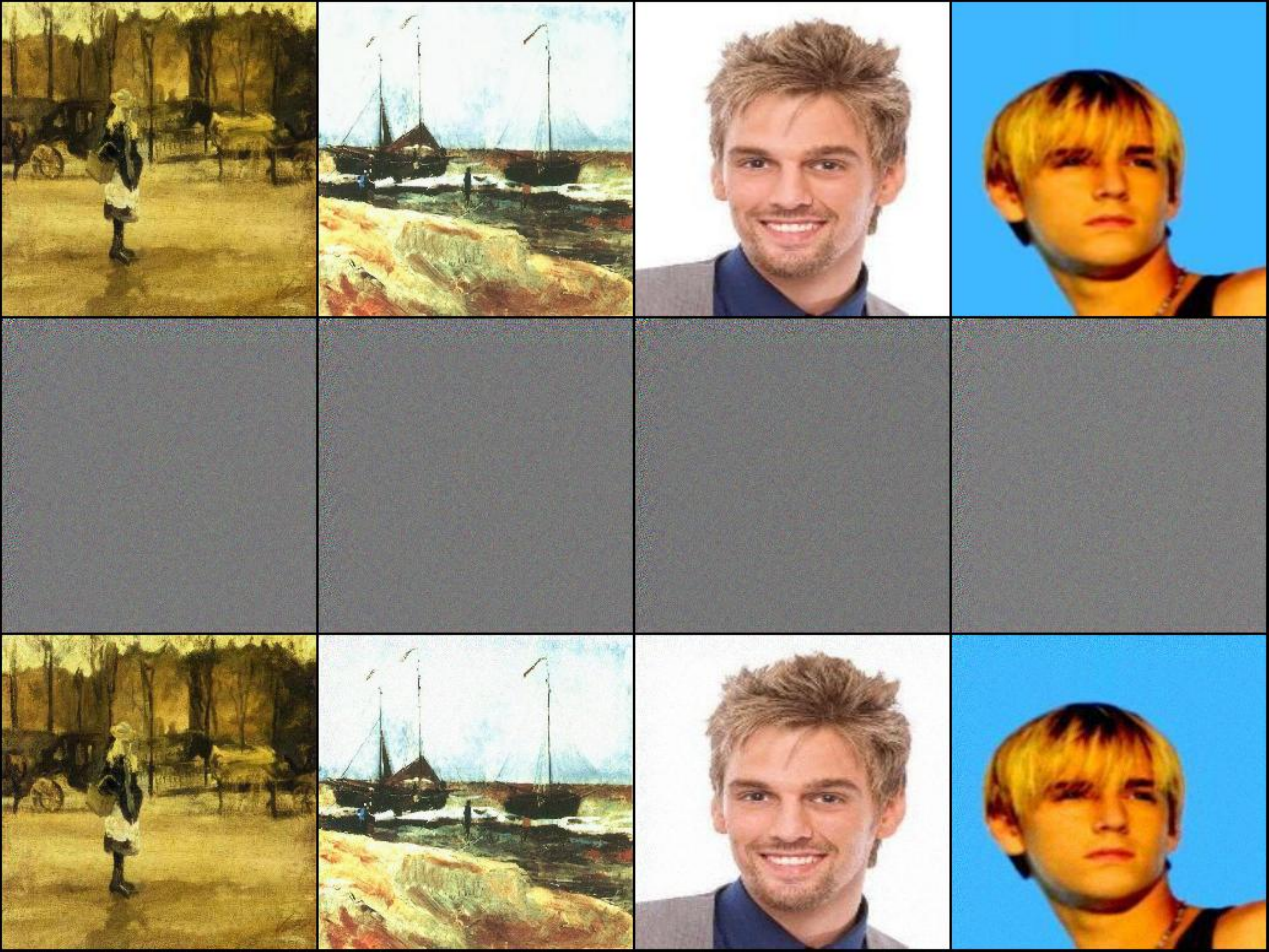} 
\caption{Clean inputs (top), added watermarks (middle), and the resulting watermarked inputs (bottom). Watermarks are multiplied by 10 for better visualization.}
\label{figure:watermark}
\end{figure} 

\begin{figure*}[!t]
\centering
\includegraphics[width=\textwidth]{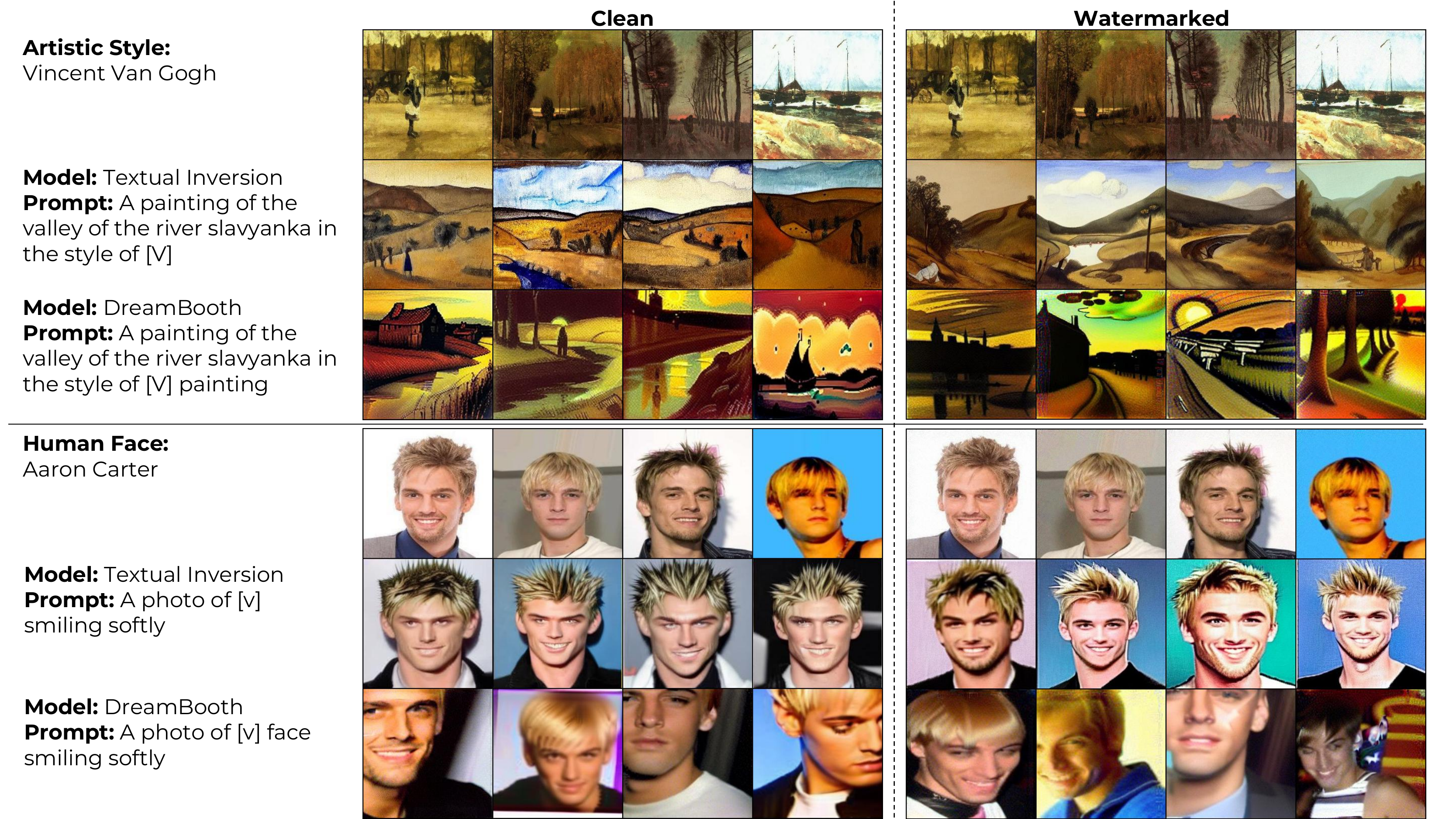} 
\caption{Images synthesized based on clean (left) vs. watermarked (right) for artistic style (top) and human face (bottom). Additional examples for other subjects with different prompts can be found in \autoref{figure:anotherexample} of \autoref{section:appendixa}.}
\label{figure:example2}
\end{figure*}

\mypara{Watermark Uniqueness}
An important property of the watermarking technology is to ensure that the watermark is unique to each individual subject.
To measure such uniqueness in our case, we measure the classification accuracy of a detector designed for a specific subject in distinguishing its own watermarked images from those generated for other subjects.
Specifically, for the artistic style task, we use the detector that is fine-tuned for Vincent Van Gogh to distinguish its own watermarked synthesized images from those of the other 26 artists.
For the human face task, we choose Aaron Carter vs. the other 3 celebrities.
We ensure the class balance and test 250 images for each.
As can be seen from \autoref{tab:cross_subject}, for the artistic style task, the classification accuracy is 84.3\% averaged on two models, which implies the high uniqueness of our \MethodName.
The result for the human face task is lower, i.e., 77.4\%.
However, we should note that in this task, uniqueness is not practically meaningful since a person only has authority over their own face and so would not test an image that contains others' faces.

\begin{table}[!t]
\centering
\caption{Watermark uniqueness measured by the classification accuracy (\%) of a detector designed for a specific subject in distinguishing its own watermarked images from those generated for other subjects.}
\label{tab:cross_subject}
\begin{tabular}{cc|cc}
\toprule
\multicolumn{2}{c|}{Artistic style}& \multicolumn{2}{c}{Human face}  \\
\midrule
TI                     & DB                    & TI                  & DB                 \\
\midrule
83.2               & 85.4                     & 76.5                   & 78.2                \\
\bottomrule
\end{tabular}
\end{table}

\subsection{Results on Image Synthesis Quality}

In addition to the detection accuracy, a successful watermark should also have little impact on the model utility, i.e., the quality of the image synthesis.
Here we evaluate how adding watermarks to the input images would affect the original synthesis quality with clean inputs.
Specifically, we measure how the FID score calculated between synthesized images and input images would change after injecting the watermarks.
As can be seen from \autoref{figure:fid}, the FID score changes only by less than 1\% on average, indicating that injecting watermarks indeed has little impact on the original synthesis quality.
In addition, we can see that Textual Inversion leads to consistently lower FID scores than DreamBooth, suggesting its better performance in subject-driven image synthesis.

\mypara{Watermark Invisibility}
\autoref{figure:watermark} visualizes examples of clean vs. watermarked input images and their corresponding watermarks.
We can observe that the watermarks in the watermarked images are barely visible to the human eye, especially for the artistic style task involving rich image textures.
Moreover, as can be seen from \autoref{figure:example2}, there are also no obvious patterns left in the output images that are synthesized based on the watermarked inputs, demonstrating that the watermarks barely affect the utility of image synthesizing quality.

\mypara{Qualitative Analyses}
We qualitatively compare the image examples synthesized using clean vs. watermarked in \autoref{figure:example2}. 
In general, we can find that in both the artistic style and human face tasks, Textual Inversion and DreamBooth models can synthesize images that are similar to the original (clean) inputs.
For the artistic style task, we choose artwork from Vincent Van Gogh, one of the most famous and influential painters in Western art history.
As can be seen, the synthesized images based on the watermarked inputs share a very similar style with those based on the clean inputs.
This is consistent with our FID results reported in \autoref{figure:fid}. 

For the human face task, we choose pictures from Aaron Carter, an American singer/rapper.
Similar to the above observation, the synthesized images based on the watermarked inputs also well maintain the face and follow a similar style to those based on the clean inputs.
An interesting finding here is that Textual Inversion and DreamBooth synthesize images with different preferences.
Specifically, DreamBooth synthesizes more realistic faces in novel views, while Textual Inversion images look more like computer-generated graphics with a fixed front and comic-like view.

\subsection{Ablation Studies}
\label{section:ablation}

\begin{figure}[!t]
    \centering
    \includegraphics[width=\columnwidth]{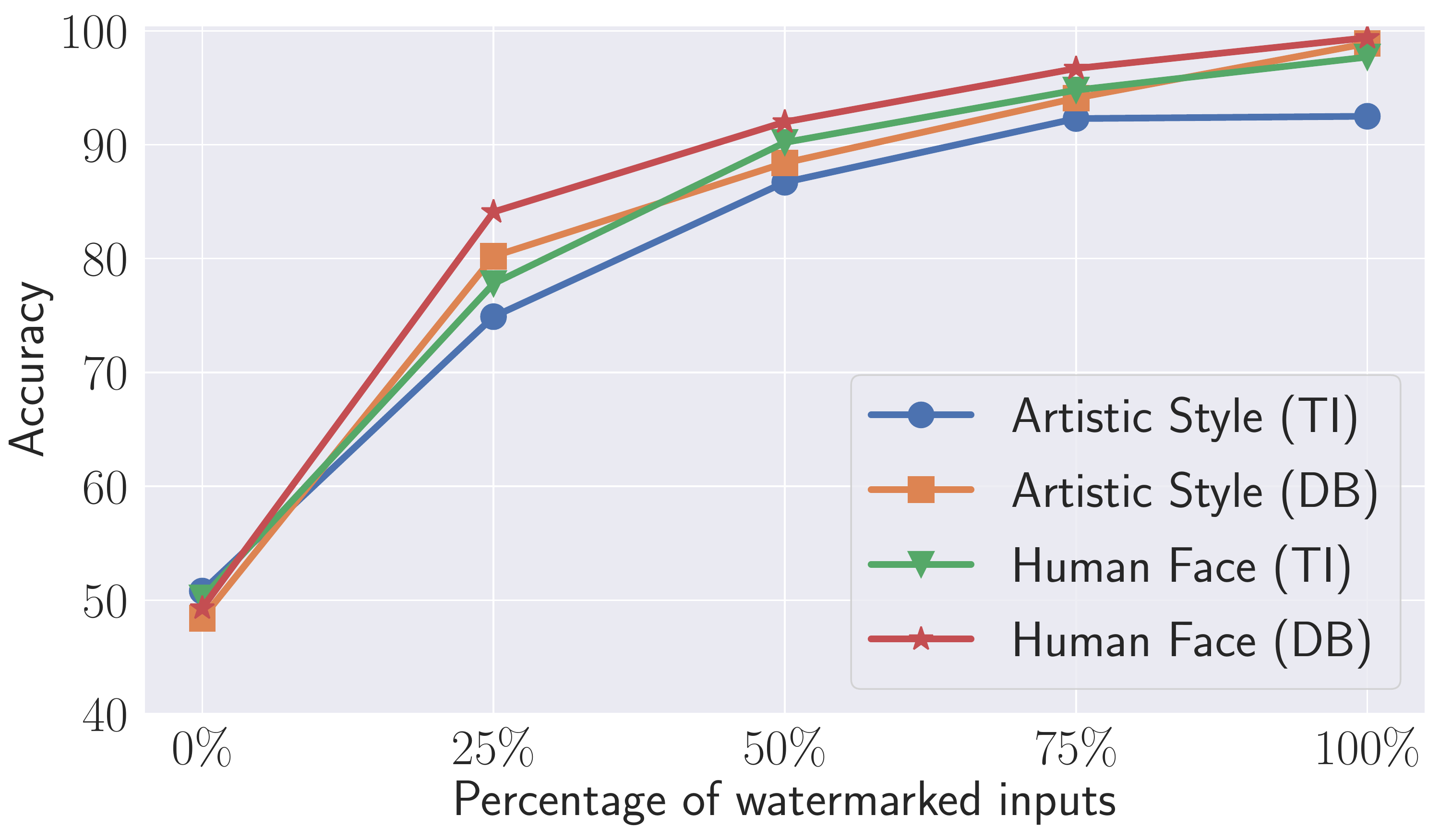}
    \caption{Watermark detection accuracy (\%) when only partial input images are watermarked.}
    \label{figure:robustness1}
\end{figure}

\mypara{Partial Input Watermarking}
In real-world scenarios, the subject owners may have already exposed their clean images before the watermark technique is developed, so it is possible for the malicious subject synthesizers to mix these images with watermarked images when training the subject-driven model.
Here we evaluate the impact of such a realistic scenario on the detection performance of our \MethodName.
\autoref{figure:robustness1} reports the performance of \MethodName when only partial input images are watermarked, in Scenario 1.

We can see that the detection accuracy of \MethodName gradually decreases as the subject synthesizers have access to more clean images.
This is expected since it becomes harder for the synthesis models to learn the watermark information when it only appears in fewer input images.
However, even when only 25\% of the input images are watermarked, \MethodName is still effective with a detection accuracy of about 80\% on average.
The accuracy increases to about 90\% when only half of the input images are watermarked.
This validates the robustness of our \MethodName against the partially watermarked situation.

\begin{figure}[!t]
    \centering
    \includegraphics[width=\columnwidth]{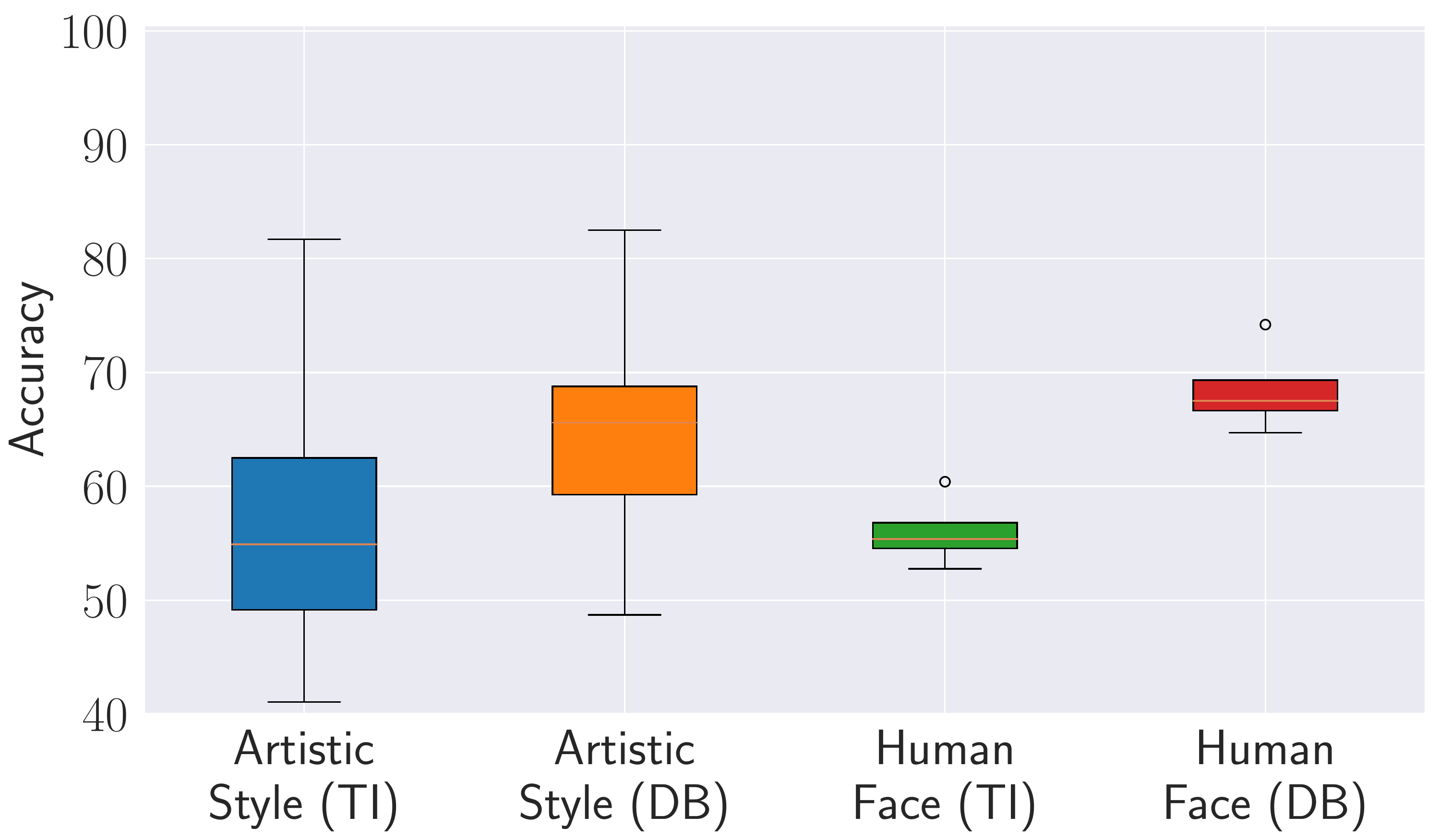} 
    \caption{Watermark detection accuracy (\%) without fine-tuning the detector.}
    \label{figure:w/o_finetuning}
    \vspace{-0.2cm}
\end{figure}

\mypara{Without Detector Fine-Tuning}
There is a detector fine-tuning phase during training \MethodName.
As discussed, this phase aims to improve the detection by ``personalizing'' the detector for each individual subject owner.
\autoref{figure:w/o_finetuning} shows the detection results when this fine-tuning phase is not applied during training \MethodName.
Note that only the results for Scenario 1 are reported since the cross-model and cross-prompt transfer scenarios only apply to the fine-tuning phase.
We can see that the detection accuracy substantially decreases compared to those corresponding results in \autoref{figure:meanaccuracy}, demonstrating the necessity of our detector fine-tuning.

\mypara{Hyperparameters $p$ and $\alpha$}
Here we analyze the impact of two main hyperparameters, i.e., the invisibility level $p$ and balancing factor $\alpha$, on the performance of our \MethodName.

\begin{figure}[!t]
\centering
\includegraphics[width=\columnwidth]{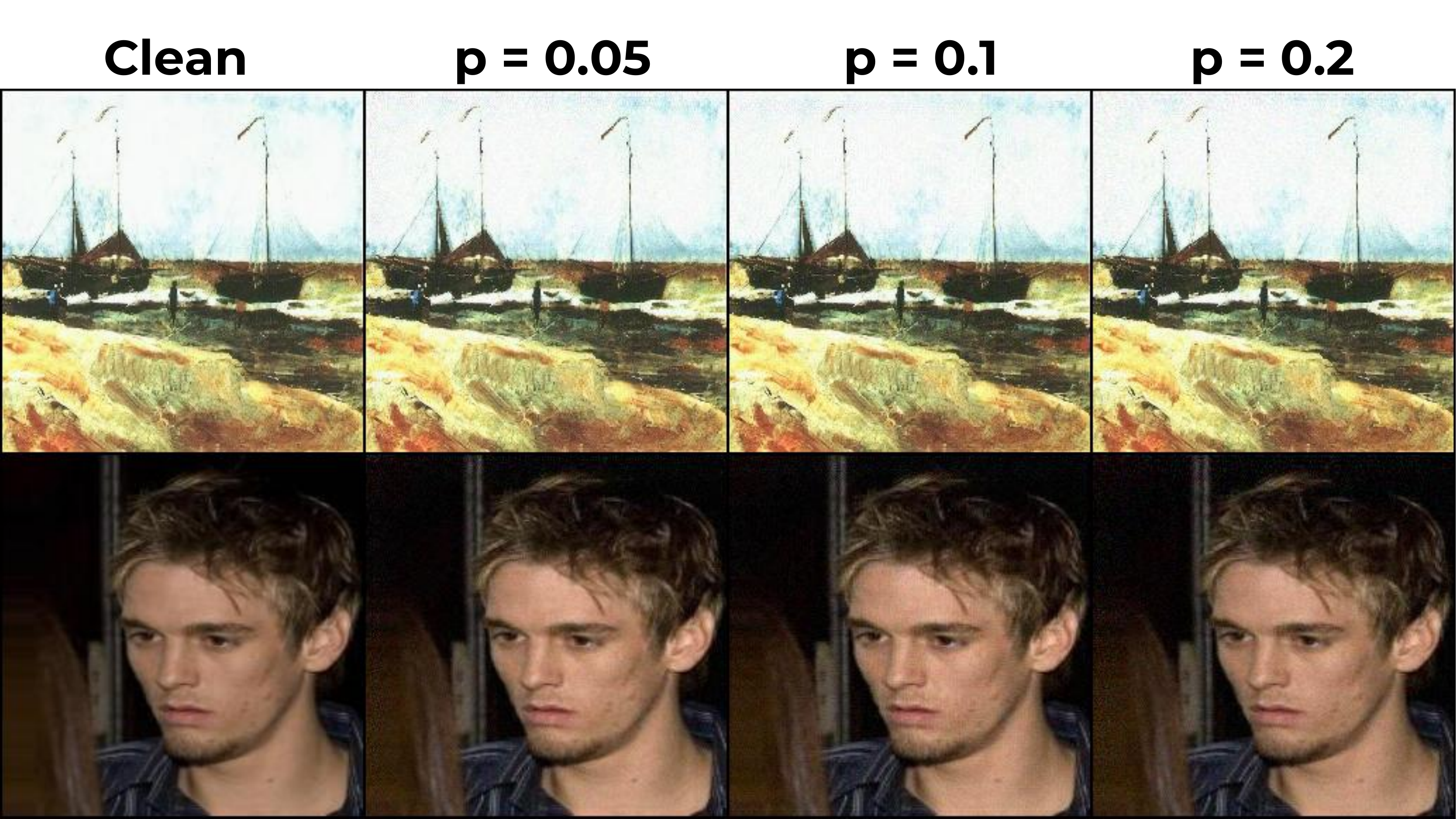} 
\caption{Clean and watermarked images under varied invisibility level $p$.}
\label{figure:example_p}
\end{figure}

\begin{figure}[!t]
\centering
\includegraphics[width=0.495\columnwidth]{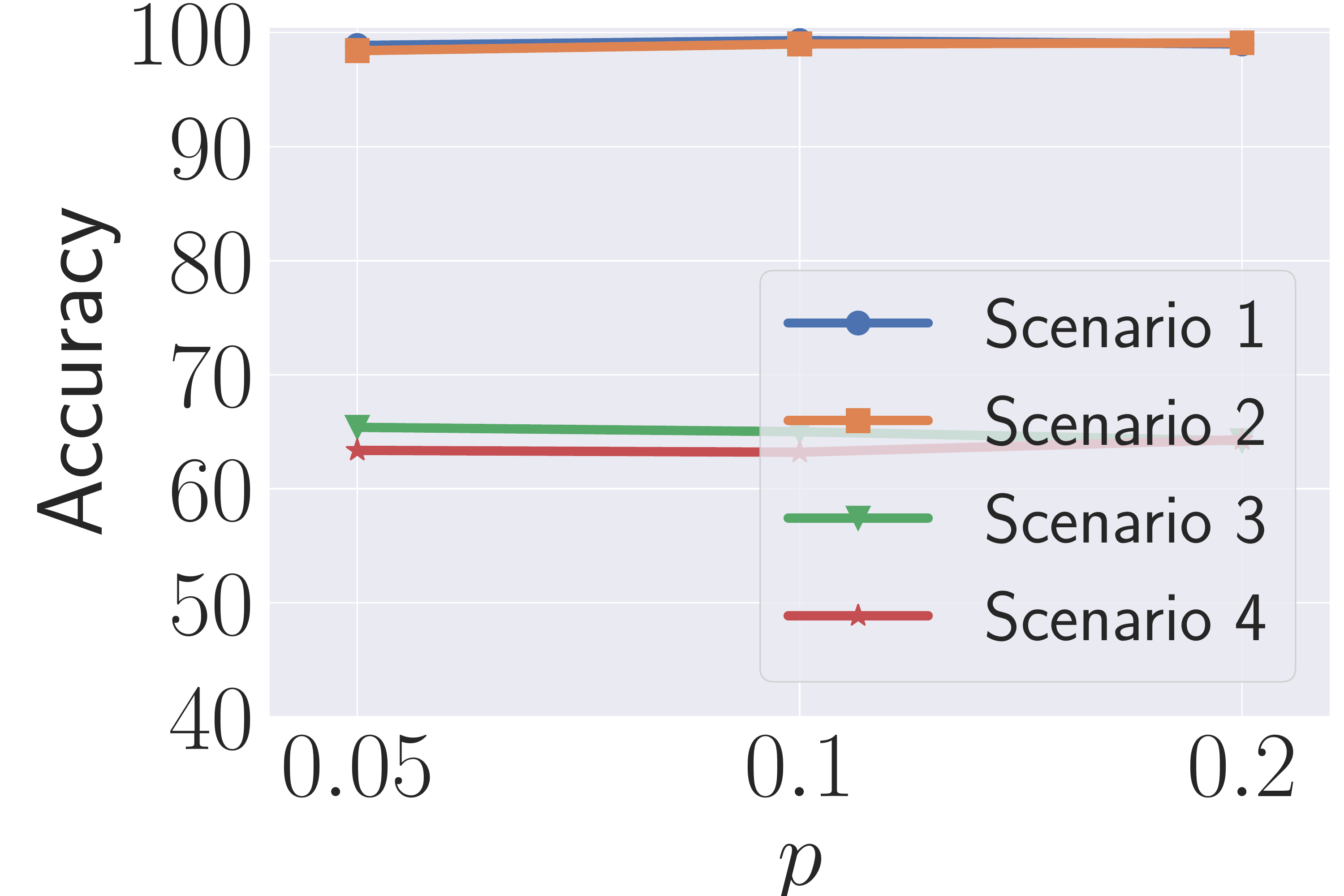}
\includegraphics[width=0.495\columnwidth]{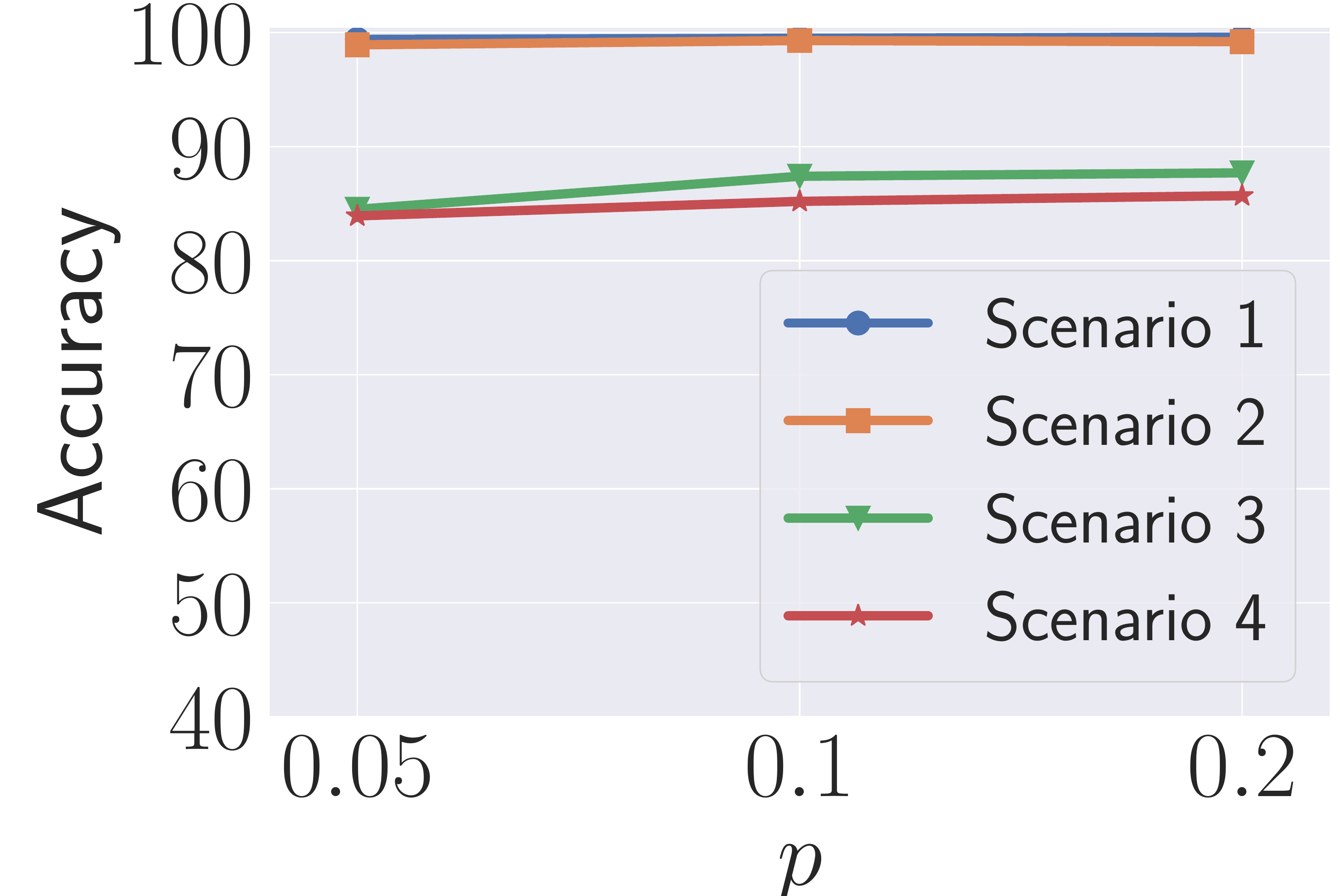}
\includegraphics[width=0.495\columnwidth]{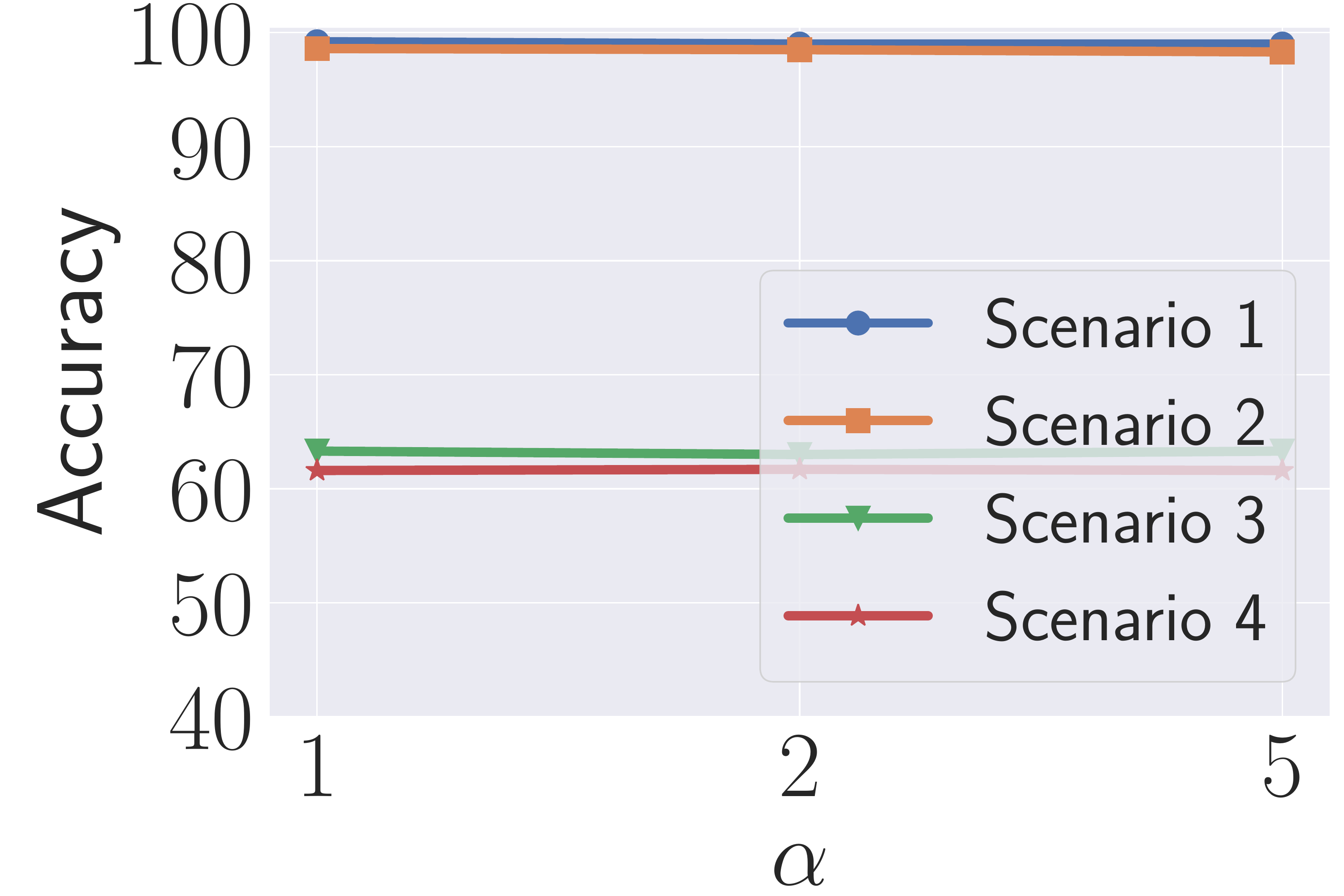}
\includegraphics[width=0.495\columnwidth]{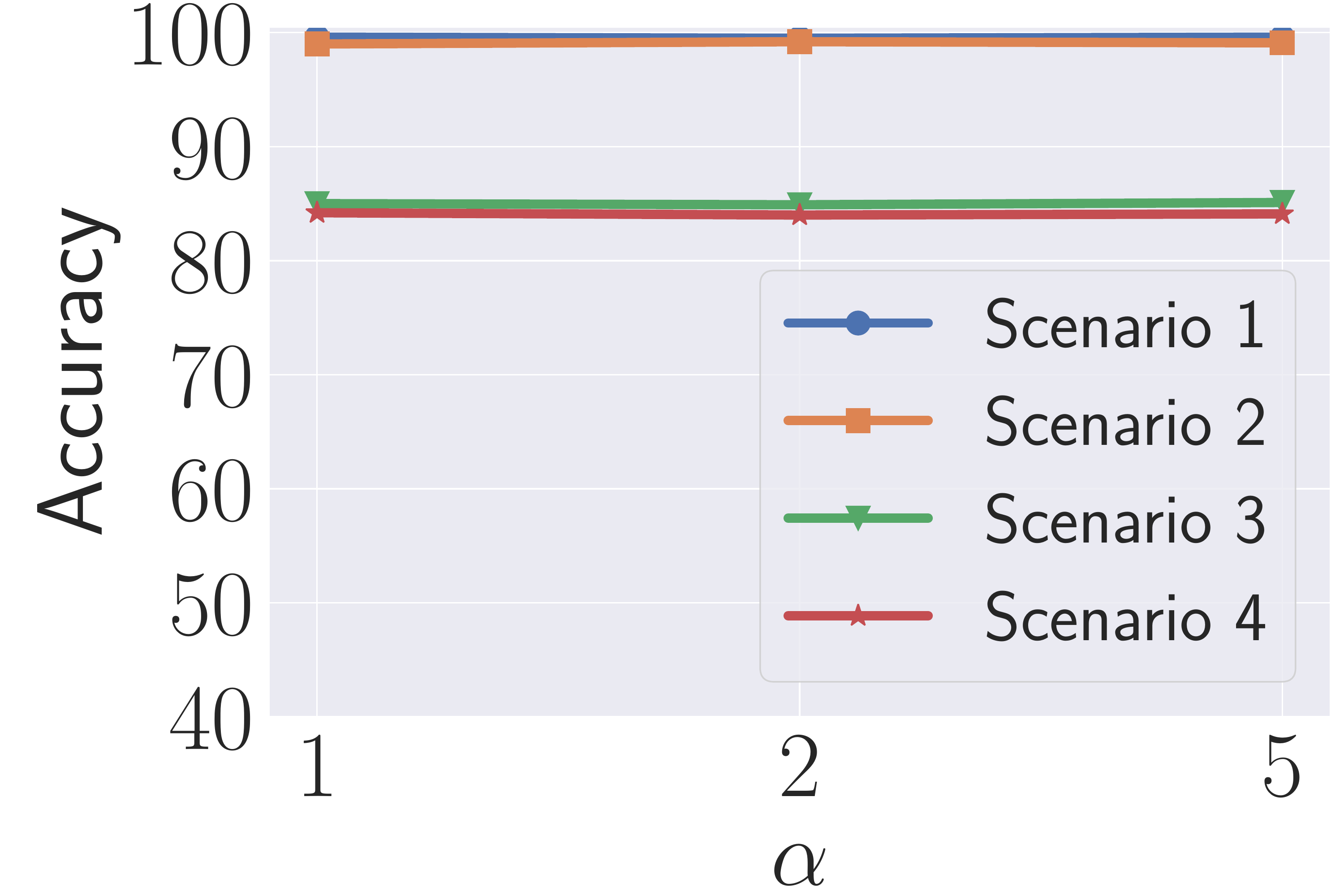}
\caption{Watermark detection accuracy (\%) under varied invisibility level $p$ and balancing factor $\alpha$ for the artistic style (left) and human face (right) tasks. DreamBooth is used as the target model, and the results for Textual Inversion with a similar pattern can be found in \autoref{figure:p_2} of \autoref{section:appendixa}.}
\label{figure:p}
\end{figure}

\autoref{figure:example_p} visualizes examples of the clean images and watermarked images under varied invisibility levels $p$.
As expected, the noise becomes more visible as $p$ is increased.
For example, $p=0.1$ leads to clearly more noticeable noise than that with $p=0.05$.
In addition, $p$ also has little impact on the detection accuracy of \MethodName, as shown in \autoref{figure:p} (top).
Specifically, the detection accuracy increases when $p$ increases from 0.05 to 0.1, especially in Scenario 3 and Scenario 4.
This is expected since a more visible watermark is easier to be detected.
However, the detection accuracy remains almost unchanged when we further increase $p$ to 0.2.
We find that it is due to the fact that when pre-training our \MethodName, the detector has already converged without pushing the watermark to reach $p=0.2$.

\autoref{figure:p} (bottom) shows the impact of the balancing factor $\alpha$ on the performance of \MethodName.
Here we report the results for DreamBooth as the target model.
We can see that, in general, $\alpha$ has little impact on detection accuracy.
The reason might be that the detector can achieve almost 100\% accuracy in the validation dataset when fine-tuning.
However, we find that increasing $\alpha$ accelerates the convergence of the generator.
For both $p$ and $\alpha$, we find that Textual Inversion leads to very similar results, as shown in \autoref{figure:p_2} of \autoref{section:appendixa}.

\section{Countermeasures}
\label{section:counter}

In practice, the malicious subject synthesizers may apply countermeasures to reduce the protecting effects of our \MethodName.
In this section, we evaluate two potential countermeasures from different perspectives on the full datasets.

\mypara{Watermark Forgery}
Since our watermarks are basically high-frequency additive noises, as visualized in \autoref{figure:watermark}, a potential countermeasure could be that an adversary deliberately adds similar noises into synthesized images as forged watermarks such that our detector would misclassify them as watermarked images.
We evaluate if \MethodName is potentially robust to this countermeasure, and we specifically test with Gaussian noises.
\autoref{table:misclassification} shows the watermark detection accuracy when Gaussian noises are added to clean inputs or outputs that are synthesized based on clean inputs.
We find that on average, our watermark detector achieves a high accuracy of about 78\%, indicating that it is hard to forge watermarks with random noises.

\begin{table}[!t]
\centering
\caption{Watermark detection accuracy (\%) against watermark forgery.}
\label{table:misclassification}
\begin{tabular}{cc|cc}
\toprule
\multicolumn{2}{c|}{Artistic style}& \multicolumn{2}{c}{Human face}  \\
\midrule
TI                     & DB                    & TI                  & DB                 \\
\midrule
77.9               & 85.1                     & 82.2                   & 85.1                 \\
\bottomrule
\end{tabular}
\end{table}
\begin{table}[!t]
\centering
\caption{Watermark detection accuracy (\%) against watermark removal.}
\label{table:countermeasure}
\resizebox{\columnwidth}{!}{
\begin{tabular}{l|cc|cc}
\toprule
  \multirow{2}{*}{Image transformations}  & \multicolumn{2}{c|}{Artistic style} & \multicolumn{2}{c}{Human face}\\
 \cline{2-5} 
         & TI                     & DB                     & TI                  & DB                 \\ \midrule
Gaussian noise (Input)   & 76.0               & 78.0            & 79.1                   & 81.2                  \\ 
Gaussian noise (Output)   & 73.5                & 75.2            & 74.7                   & 77.3                   \\ 
\midrule
JPEG compression (Input) & 71.5                & 72.4              & 75.5                  & 77.4                    \\
JPEG compression (Output) & 64.7               & 67.2              & 69.3                  & 72.4                    \\
\bottomrule
\end{tabular}
}    
\end{table}

\mypara{Watermark Removal} Existing work has shown that adversarial perturbations are vulnerable to image transformations in both the training~\cite{LZL23} and testing~\cite{XEQ18} stages. 
Related studies in protecting against image synthesis have also considered input transformations as a typical countermeasure~\cite{YSAF21, SCWZHZ23, LWHZXSXMG23, LPNDTT23}.
Following them, we test two popular image transformations: Gaussian noise and JPEG compression.
Specifically, the malicious subject synthesizers may apply such transformations to remove the watermarks from either input or output (synthesized) images.

\autoref{figure:example_countermeasure} in \autoref{section:appendixa} visualizes the images transformed by Gaussian noise with different standard deviations and JPEG compression with different quality factors.
As can be seen, too severe transformations lead to obvious visual artifacts.
Therefore, we choose Gaussian noise with a standard deviation of 0.0005 and JPEG compression with a factor of 20 for the following experiment.

\begin{figure}
    \centering
    \includegraphics[width=\columnwidth]{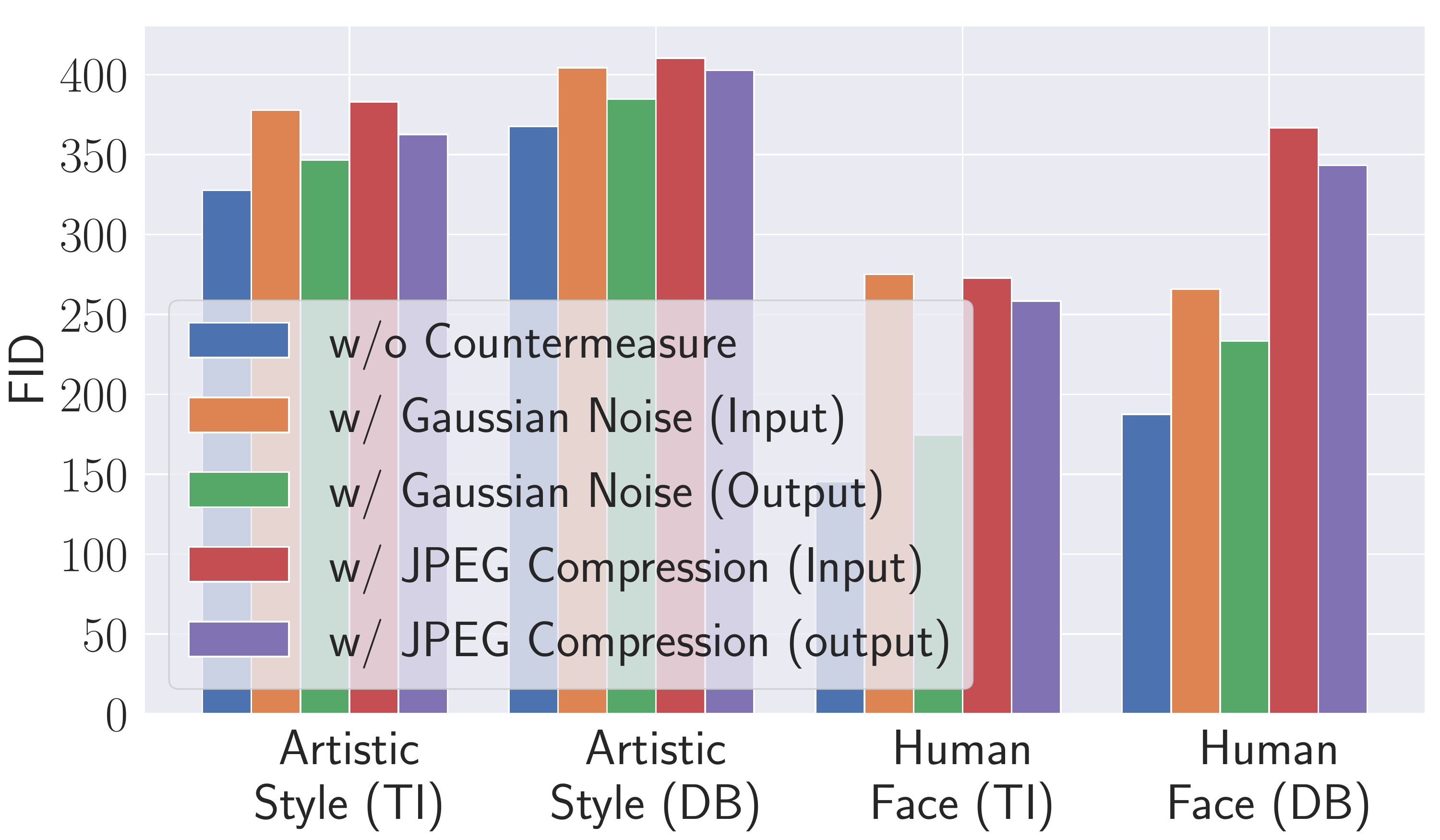}
    \caption{FID scores for synthesis from clean inputs, watermarked inputs, watermarked inputs with Gaussian noise, watermarked inputs with JPEG compression. The FID is calculated between the clean inputs and synthesized outputs.}
    \label{figure:fid_countermeasure}
\end{figure}

\autoref{table:countermeasure} shows the detection results against two image transformations and \autoref{figure:fid_countermeasure} shows their corresponding FID scores.
In general, the average detection accuracy remains high, i.e., 74.1\%.
On the other hand, the FID results show that these transformations inevitably degrade the synthesis quality.
Interestingly, transforming outputs leads to a consistently stronger countermeasure than transforming inputs as well as smaller quality degradation.
When comparing two different transformations, we can observe that although JPEG compression is stronger against our detection than Gaussian noise, it leads to greater quality degradation. 

\begin{figure*}[!t]
\centering
\includegraphics[width=\textwidth]{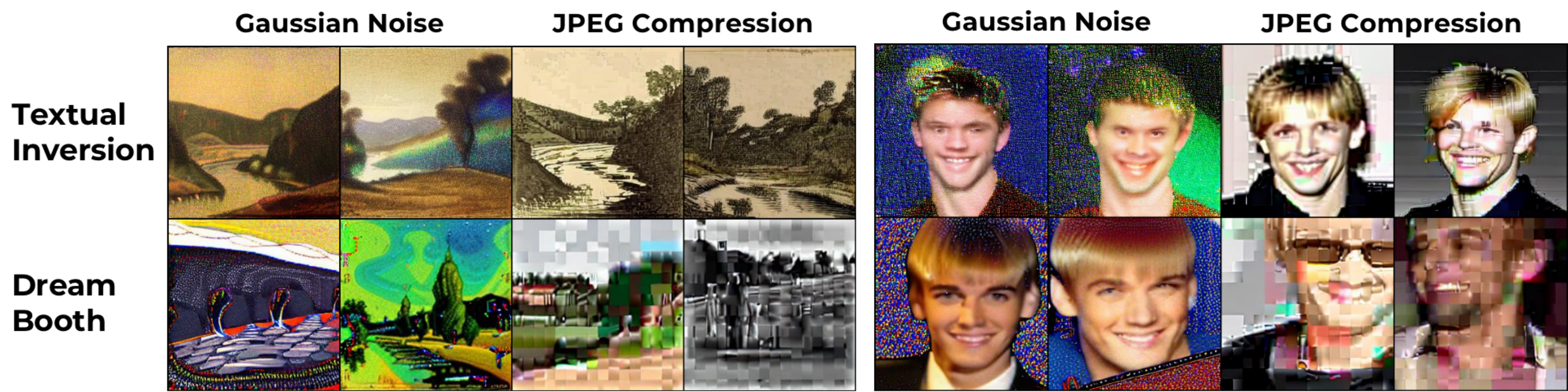} 
\caption{The effects of input transformations on the synthesized outputs. The same subjects as in \autoref{figure:example2} are used.}
\label{figure:example_countermeasureinfer}
\end{figure*}

Examples of synthesized images in \autoref{figure:example_countermeasureinfer} explain the above findings.
Specifically, we find that when transforming the inputs, the visual artifacts are magnified in the synthesized outputs.
In addition, the Gaussian noise results in high-frequency noisy patterns while the JPEG compression results in obvious blocking artifacts.
In particular, for the human face task, Textual Inversion even fails to learn the accurate facial attributes of the subject.
Besides, since DreamBooth tends to inherit fine-grained information from the input images (as discussed in \autoref{section:result_acc}), it causes severe blocking artifacts, and makes it even hard to recognize the actual image content.

Compared to the artistic style task, the human face task yields greater quality degradation.
This might be explained by our previous finding from \autoref{figure:s1_artist} that the subject-driven model trained on face images focuses more on the local attributes, and as a result, it is not robust to global transformations.
In contrast, the style models are better at capturing holistic features, so they are more robust to global transformations.
Overall, the above observations suggest that neither the watermark forgery nor watermark removal is effective in countering our \MethodName.

\section{Discussion and Limitations}
Our \MethodName is the first approach to watermarking images against unauthorized subject-driven image synthesis.
Although we have conducted extensive experiments to demonstrate the effectiveness of \MethodName in various scenarios, there is still future work to be done to further improve it.
Here we identify two potential limitations of the current \MethodName and discuss possible ways to address them.

\mypara{Cross-Model Transferability}
Cross-model transferability of \MethodName is important in practical scenarios since the target model an adversary would choose is usually not known to the image owner who wants to watermark their images.  
This might not be a big problem at this moment given that there are only a few (publically-available) recipes for training subject-driven synthesis models, e.g., Textual Inversion and DreamBooth adopted in our experiments.
However, with the rapid development of subject-driven synthesis, we expect more effective models to be proposed.

Our experiments have shown that the cross-model transferability of \MethodName is about 20\% lower than the case with a known model.
We hypothesize that this is due to the differences between Textual Inversion and DreamBooth in image synthesis.
For example, in the artistic style task, DreamBooth tends to capture fine-grained concepts from the input images while Textual Inversion focuses more on the holistic style.
In the human face task, DreamBooth tends to produce realistic images but Textual Inversion produces images like virtual characters.

Related work on adversarial examples has extensively studied the transferability of perturbations across different CNN models or other families of architectures~\cite{NRKKP22, FZWWL22, ZZLSAB22}, e.g., Vision Transformers~\cite{DBKWZUDMHGUH21}.
A typical idea is to incorporate the specific model properties. 
Similarly, we can also improve the cross-model transferability of \MethodName by incorporating model-specific properties, including our findings on their differences in synthesis, during fine-tuning the detector.

\mypara{Watermark Uniqueness} Although our \MethodName achieves substantial watermark uniqueness (e.g., 84.3\% for the artistic style task), we acknowledge that it is still not on par with conventional methods, which are based on explicit watermark injection and reconstruction (but not applicable in our scenario).
In order to improve the uniqueness, it would help if we fine-tune not only the detector but also the generator based on images from the specific subject under protection.
In this way, the property of a specific subject can be embedded into not only the detector but also the generator and as a consequence its generated watermarks.
Note that this would inevitably cost more computational resources.

\section{Conclusion and Outlook}

In this paper, we propose \MethodName, the first approach to watermarking images against subject-driven image synthesis.
Our \MethodName is essentially different from previous protections because it aims to prevent only unauthorized use of image synthesis while maintaining its utility.
\MethodName consists of a generator and a detector.
These two components are pre-trained based on large-scale data, and the detector is fine-tuned to further improve the personalized detection performance for each individual subject.
We have demonstrated the robustness and invisibility of our \MethodName in two typical image synthesis tasks, i.e., artistic style and human face, and on two representative target models, i.e., Textual Inversion and DreamBooth.
In particular, we demonstrate the generalizability of \MethodName in practical scenarios with unknown tasks, models, and text prompts, as well as with only partial data being watermarked.
Moreover, the subject-driven model trained with our watermarked inputs yields similar synthesis quality to that trained on clean inputs.

Moving forward, there are two promising directions to explore.
First, similar to all existing protections, our current \MethodName may turn out to be vulnerable to a future countermeasure, e.g., the very recent development of the perturbation purification techniques~\cite{NGHXVA22, WYG22}.
Therefore, future work should evaluate the robustness of \MethodName more comprehensively, which would in turn help design a better watermark system.  
Second, there is still room for improving \MethodName in practical scenarios, such as cross-model and partial data watermarking.
Further adapting \MethodName to specific subjects based on the characteristics of their images would potentially help. 

\newpage
\bibliographystyle{plain}
\bibliography{normal_generated_py3,additional_bib}
\newpage

\appendix
\section{Additional Tables and Figures}
\label{section:appendixa}
In this section, we list additional tables and figures.

\begin{table}
    \centering
    \caption{Artists and the corresponding genres.}
    \label{table:artists}
\resizebox{\columnwidth}{!}{
    \begin{tabular}{l|l|l}
\toprule
   Index     &  Name  & Genre\\
\midrule
    1     & Vincent Van Gogh  &  Realism \\
    2     & Claude Monet  & Impressionism \\ 
    3     & Leonardo Da Vinci  &  High renaissance \\
    4     & Pablo Picasso & Cubism \\
    5     & Agnes Martin & Minimalism \\
    6     & Aubrey Beardsley & Art nouveau modern \\
    7     & Caravaggio & Baroque \\
    8     & Edvard Munch & Expressionism \\
    9     & Edward Hopper & New realism \\
    10     & Fra Angelico & Early renaissance \\
    11     & Francisco Goya & Romanticism \\
    12     & Francois Boucher & Rococo \\
    13     & Franz Kline & Action painting\\
    14     & Georges Braque & Analytical cubism \\
    15     & Georges Seurat & Pointillism \\
    16     & Giorgio Vasari & Mannerism late renaissance \\
    17     & Gustave Moreau & Symbolism \\
    18     & Henri Rousseau & Naive art primitivism\\
    19     & Henry Matisse & Fauvism \\
    20     & Hiroshige & Ukiyo e \\
    21     & Jackson Pollock & Abstract expressionism \\
    22     & Jan Van Eyck & Northern renaissance \\
    23     & Juan Gris & Synthetic cubism \\
    24     & Mark Rothko & Color field\\
    25     & Neil Welliver & Contemporary realism \\
    26     & Paul Cezanne & Post impression \\
    27     & Roy Lichtenstein & Pop art \\
\bottomrule
    \end{tabular}
    }
\end{table}

\begin{table*}[!t]
\centering
\caption{All 30 prompts for artistic style and 30 prompts for human face in our experiments.}
\label{table:all_prompts}
\resizebox{\textwidth}{!}{
\begin{tabular}{l|l}
\toprule
Artistic Style & Human Face \\
\midrule
A painting of New York City in the style of [$\mathbf{V}$] & A photo of [$\mathbf{V}$] smiling softly \\
A painting of woman with animals in the style of [$\mathbf{V}$] & A photo of [$\mathbf{V}$] wearing a colorful men's suit \\
A painting of guitar and fruit dish in the style of [$\mathbf{V}$] & A photo of [$\mathbf{V}$] smoking on chair \\
A painting of the valley of the river slavyanka in the style of [$\mathbf{V}$] & A photo of [$\mathbf{V}$] in a fluffy pink robe \\
A painting of a girl with fruits in the style of [$\mathbf{V}$] & A photo of [$\mathbf{V}$] with straight dark hair \\
A painting of murillo boy with a dog in the style of [$\mathbf{V}$] & A photo of [$\mathbf{V}$] holding a change cup \\
A painting of a tree full of birds in the style of [$\mathbf{V}$] & A photo of [$\mathbf{V}$] with a cup of coffee \\
A painting of shaddow on Frankfork barren in the style of [$\mathbf{V}$] & A photo of [$\mathbf{V}$] in a crime scene \\
A painting of bathing woman in the style of [$\mathbf{V}$] & A photo of [$\mathbf{V}$] escaping out of a house \\
A painting of cattles in the spring in the style of [$\mathbf{V}$] & A photo of [$\mathbf{V}$] killing terrorista with a gunshot \\
A painting of pink house in the style of [$\mathbf{V}$] & A photo of [$\mathbf{V}$] with long curly hair \\
A painting of church on Lenox avenue in the style of [$\mathbf{V}$] & A photo of [$\mathbf{V}$] eating a bagel \\
A painting of study for the last supper in the style of [$\mathbf{V}$] & A photo of [$\mathbf{V}$] warning a jacket \\
A painting of view of the beach and sea from the mountains Crimea in the style of [$\mathbf{V}$] & A photo of [$\mathbf{V}$] with cat head \\
A painting of man in military costume in the style of [$\mathbf{V}$] & A photo of [$\mathbf{V}$] looking at the sky \\
A painting of cowboy in the organ mountains New Mexico in the style of [$\mathbf{V}$] & A photo of [$\mathbf{V}$] holding a tablet \\
A painting of manufactures on a blue background in the style of [$\mathbf{V}$] & A photo of [$\mathbf{V}$] sitting on couch \\
A painting of girl wit hdoll in the style of [$\mathbf{V}$] & A photo of [$\mathbf{V}$] reading newspaper \\
A painting of figures in a park Paris in the style of [$\mathbf{V}$] & A photo of [$\mathbf{V}$] holding a book \\
A painting of boy with blue cap in the style of [$\mathbf{V}$] & A photo of [$\mathbf{V}$] with cat \\
A painting of Mary and child in the style of [$\mathbf{V}$] & A photo of [$\mathbf{V}$] with a coconut \\
A painting of the young peasant and his wife in the style of [$\mathbf{V}$] & A photo of [$\mathbf{V}$] carrying water \\
A painting of young woman study for the clearing in the style of [$\mathbf{V}$] & A photo of [$\mathbf{V}$] in the jungle \\
A painting of statue of liberty in the style of [$\mathbf{V}$] & A photo of [$\mathbf{V}$] working in the hospital \\
A painting of naked woman in a landscape in the style of [$\mathbf{V}$] & A photo of [$\mathbf{V}$] in the rain \\
A painting of trees in the lake in the style of [$\mathbf{V}$] & A photo of [$\mathbf{V}$] laying in the green grass\\
A painting of sailboats in the bay in the style of [$\mathbf{V}$] & A photo of [$\mathbf{V}$] in the clinic stetic \\
A painting of guitar player in the style of [$\mathbf{V}$] & A photo of [$\mathbf{V}$] with beautiful flowers \\
A painting of mountains in the clouds in the style of [$\mathbf{V}$] & A photo of [$\mathbf{V}$] in the street\\
A painting of the reader in the style of [$\mathbf{V}$] & A photo of [$\mathbf{V}$] in middle age \\
\bottomrule
\end{tabular}
}
\end{table*}

\begin{figure*}[!t]
\centering
\includegraphics[width=0.9\textwidth]{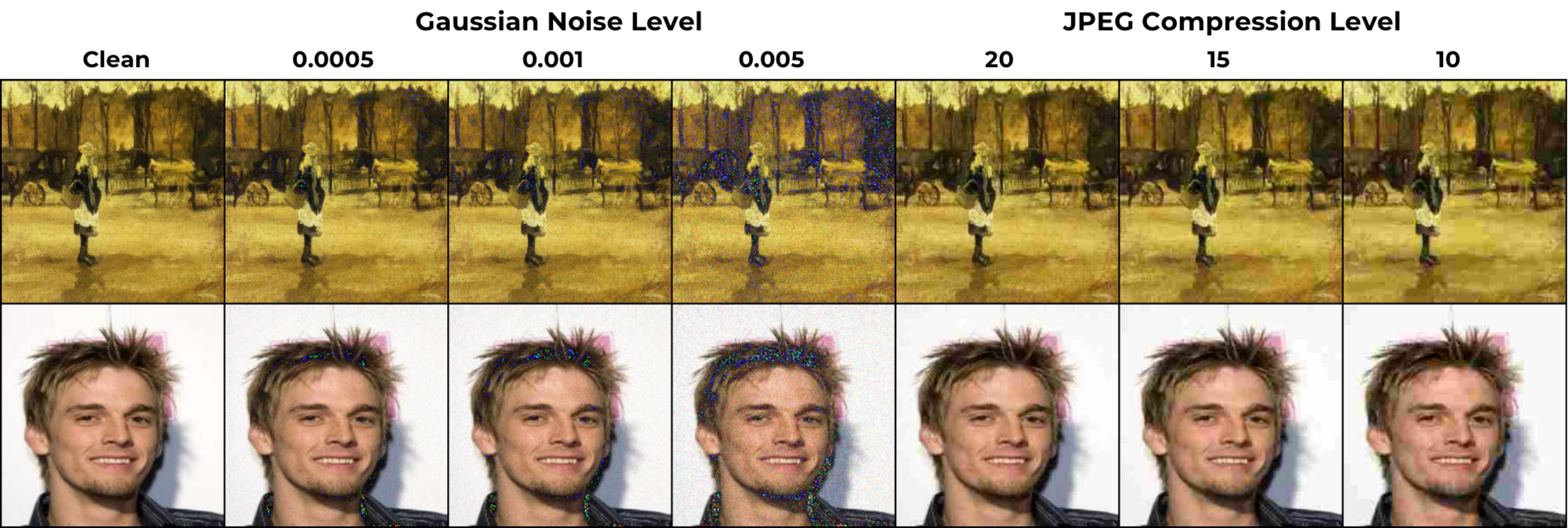} 
\caption{The effects of input transformations on the clean inputs.}
\label{figure:example_countermeasure}
\end{figure*}

\begin{figure*}[!t]
\centering
\begin{subfigure}{0.66\columnwidth}
\includegraphics[width=\columnwidth]{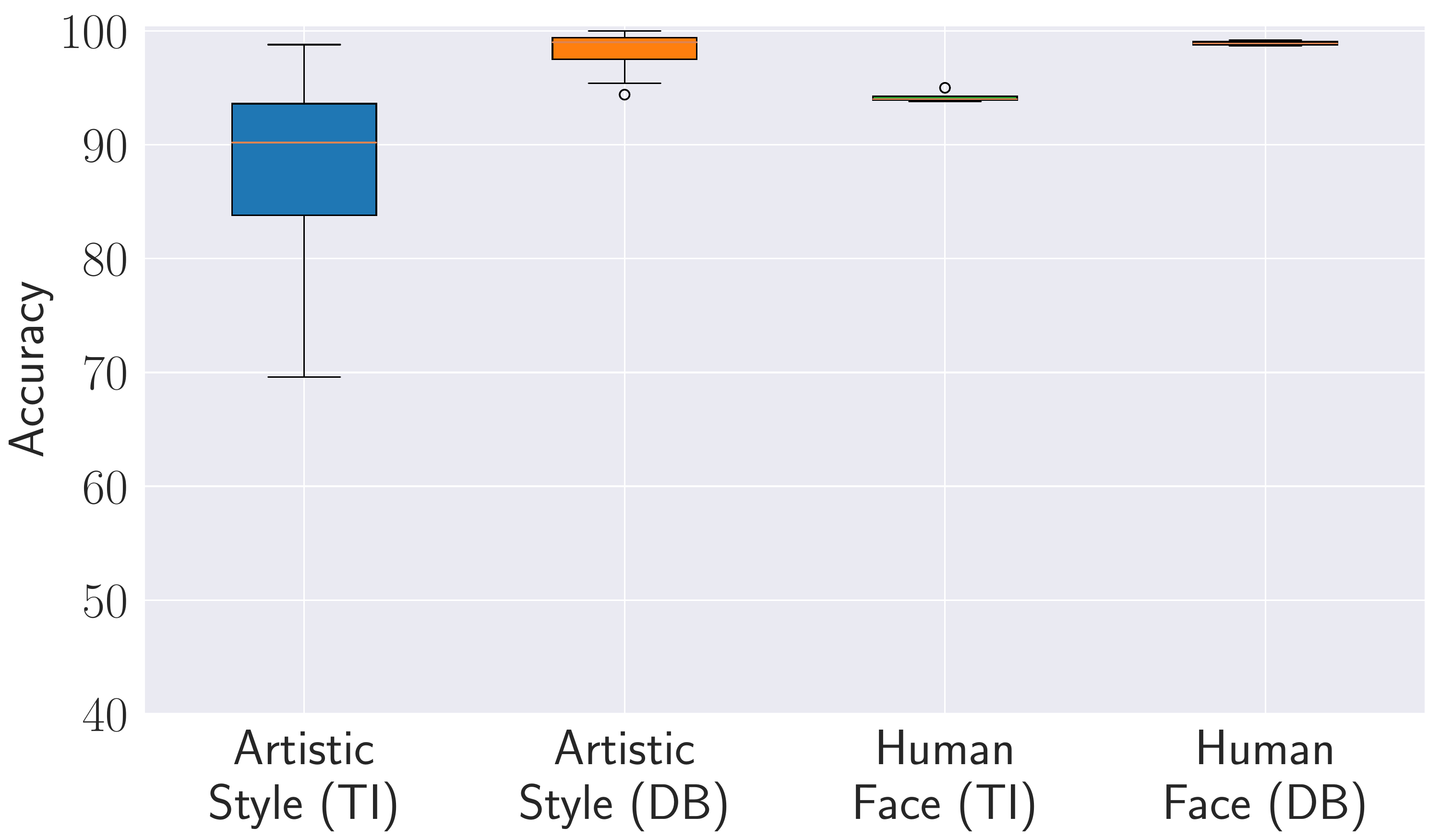}
\caption{Scenario 2}
\label{figure:s2}
\end{subfigure}
\begin{subfigure}{0.66\columnwidth}
\includegraphics[width=\columnwidth]{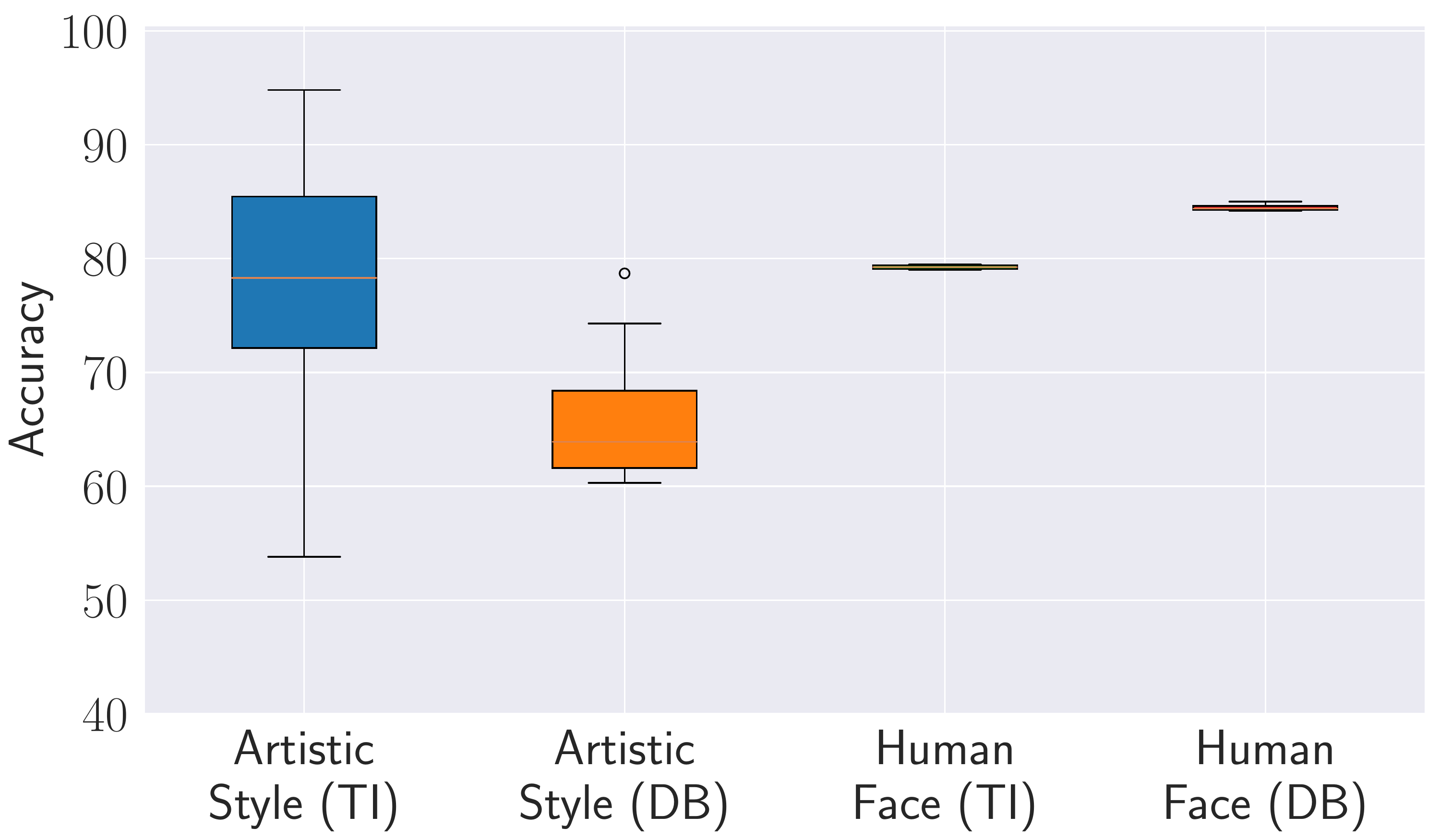}
\caption{Scenario 3}
\label{figure:s3}
\end{subfigure}
\begin{subfigure}{0.66\columnwidth}
\includegraphics[width=\columnwidth]{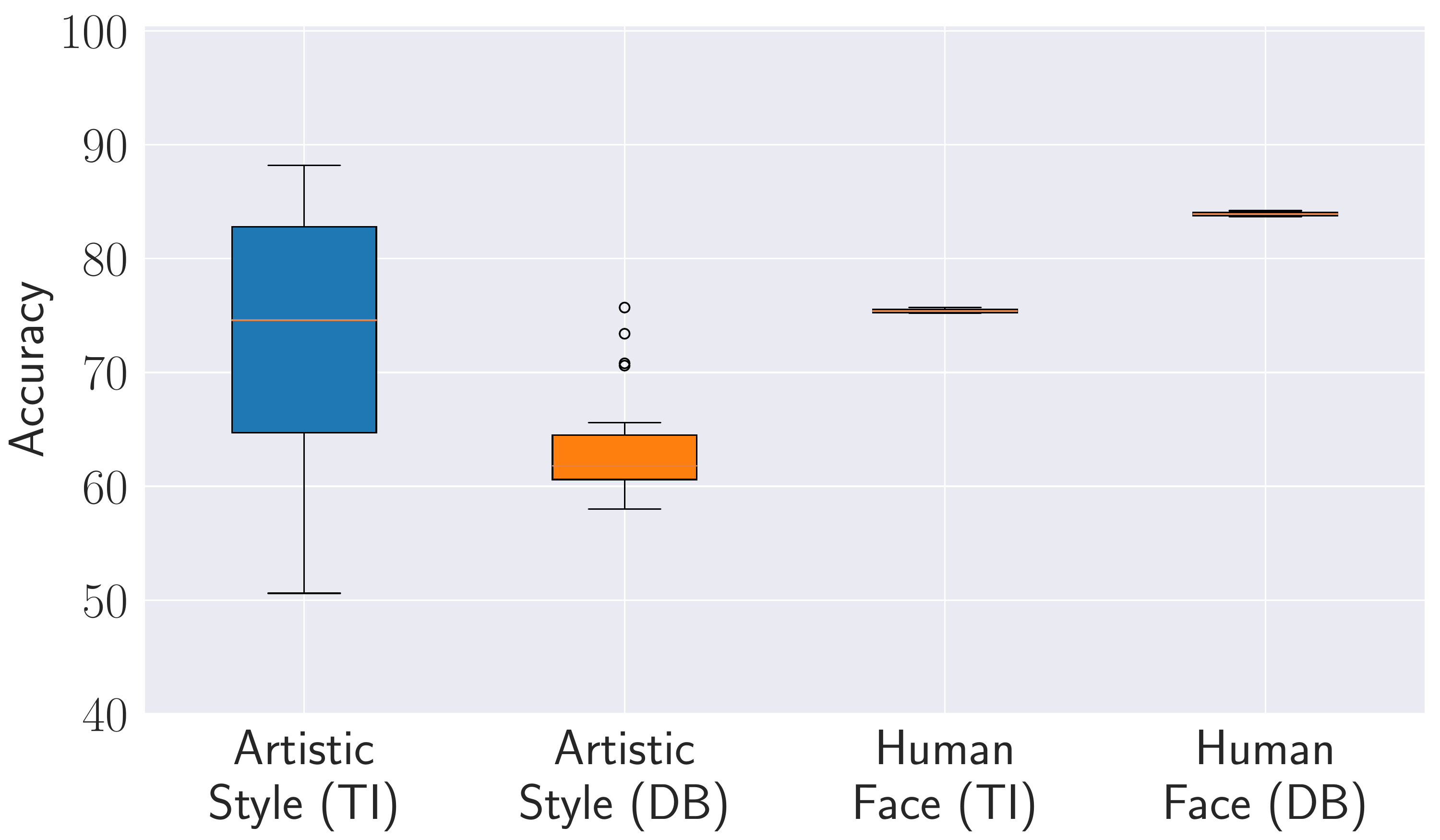}
\caption{Scenario 4}
\label{figure:s4}
\end{subfigure}
\caption{Detection accuracy of \MethodName in different scenarios, supplement to \autoref{figure:s1_artist}.}
\label{figure:s2s3s4}
\end{figure*} 

\begin{figure*}[!t]
\centering
\includegraphics[width=0.68\columnwidth]{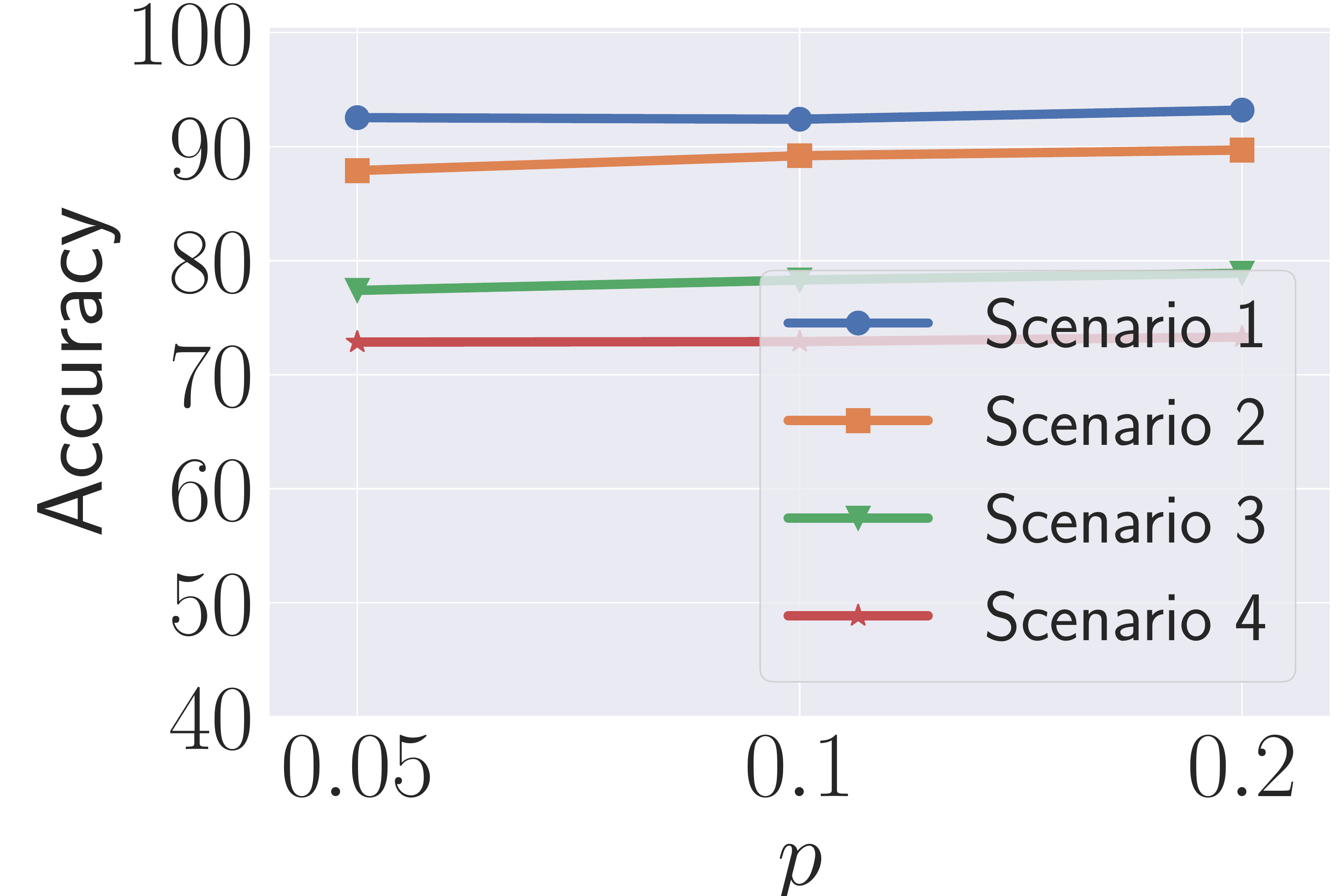}
\includegraphics[width=0.68\columnwidth]{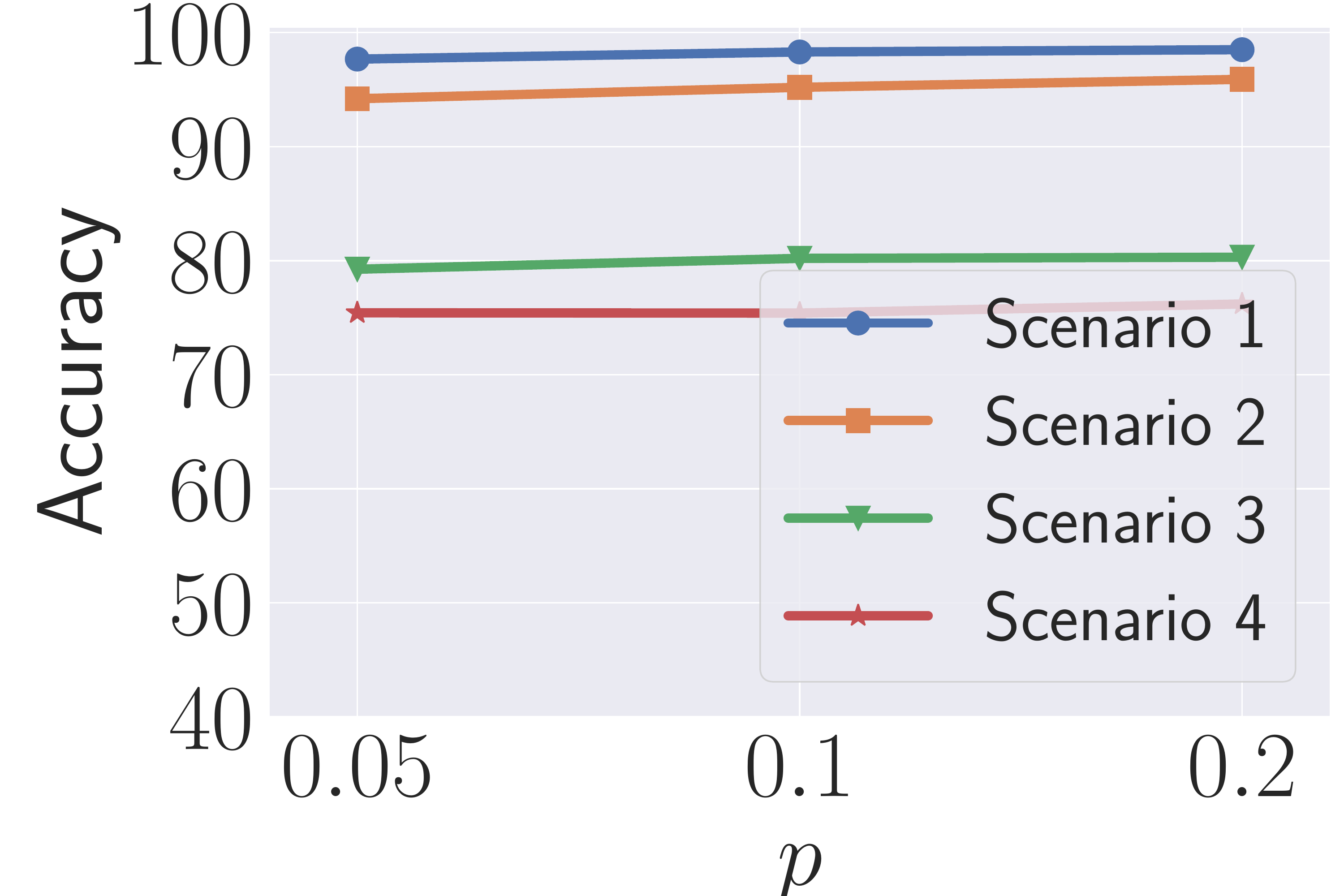}

\includegraphics[width=0.68\columnwidth]{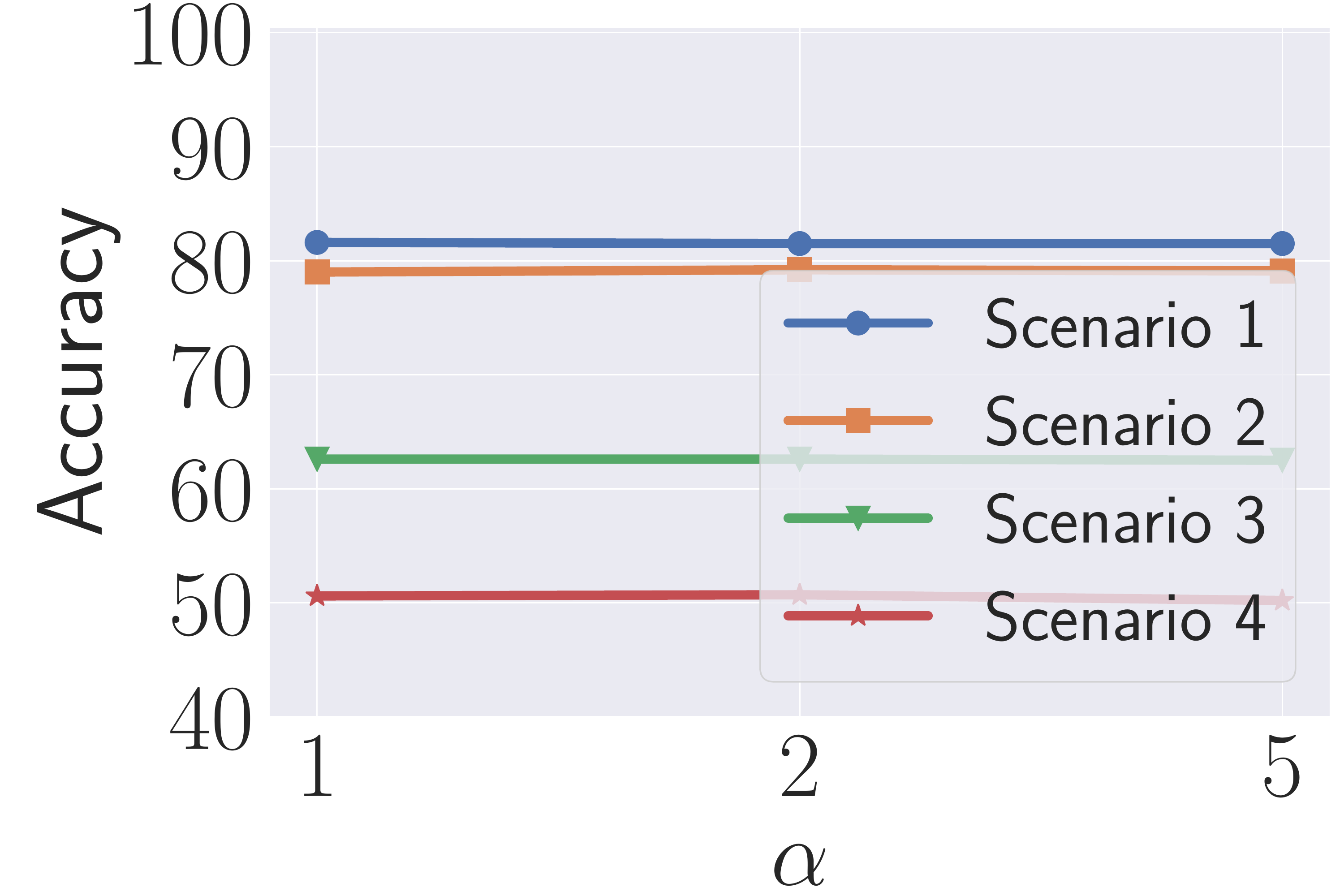}
\includegraphics[width=0.68\columnwidth]{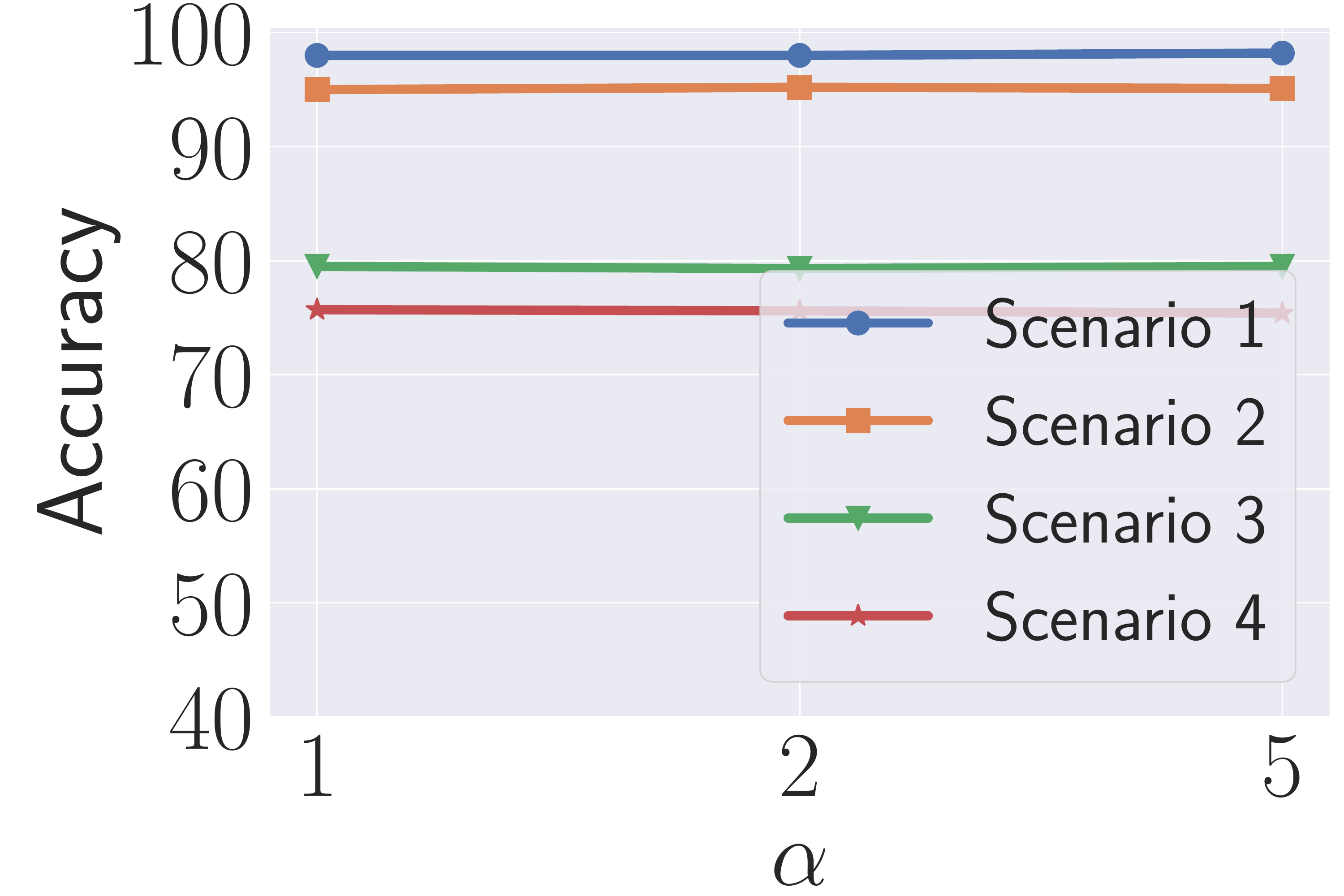}
\caption{Watermark detection accuracy (\%) under varied invisibility level $p$ (top) and balancing factor $\alpha$ (bottom) for artistic style (left) and human face (right). Textual Inversion is used as the target model.}
\label{figure:p_2}
\end{figure*} 

\begin{figure*}
    \centering
    \includegraphics[width=\textwidth]{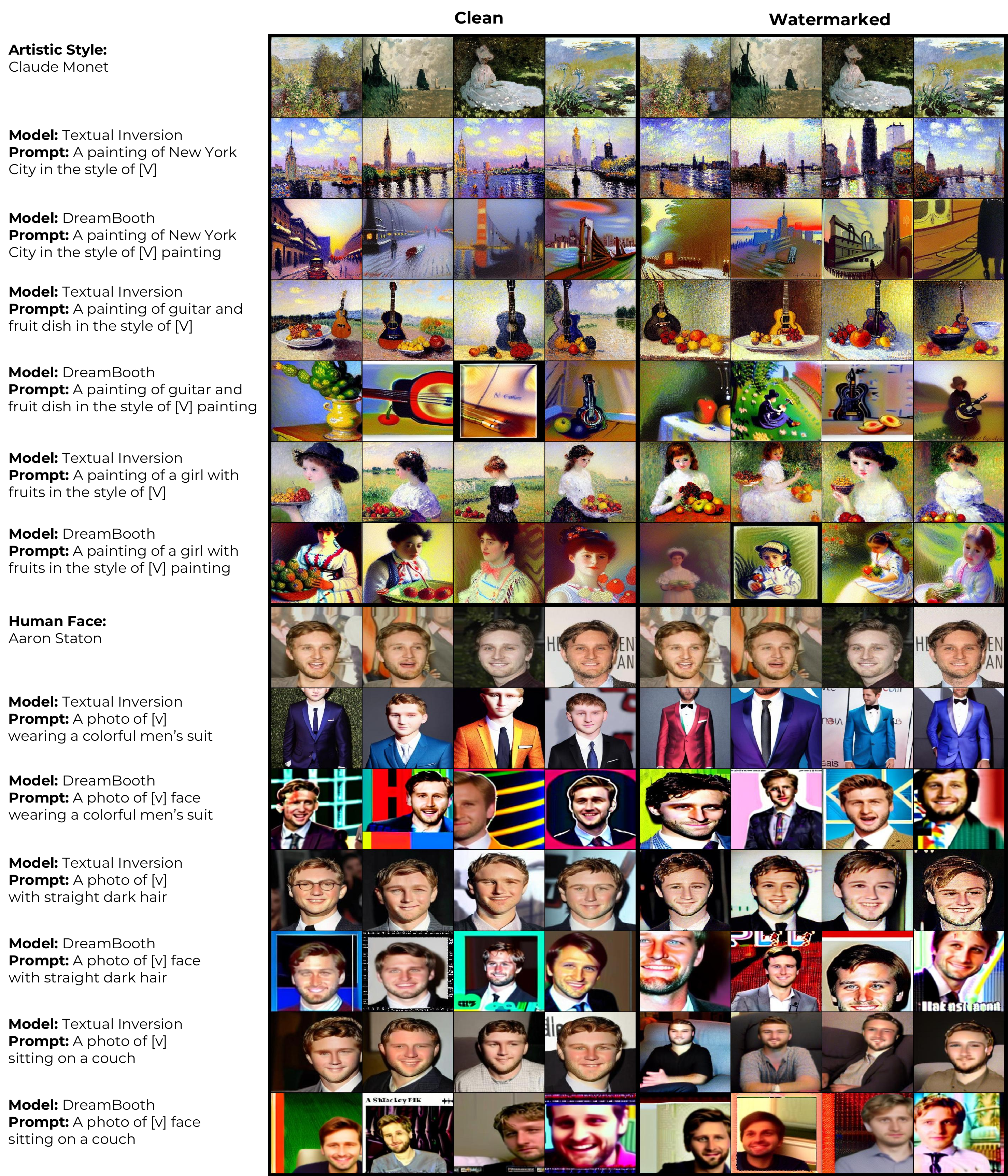}
    \caption{Images synthesized based on clean (left) vs. watermarked (right) for artistic style (top) and human face (bottom).}
    \label{figure:anotherexample}
\end{figure*}

\end{document}